\documentclass[11pt,a4paper]{article}

\usepackage[T1]{fontenc}
\usepackage{mathptmx}            
\usepackage[scaled=0.92]{helvet} 
\usepackage{courier}             

\usepackage[a4paper,top=1in,bottom=1in,left=1.15in,right=1.15in]{geometry}
\usepackage{setspace}
\usepackage[explicit]{titlesec}
\titleformat{\section}{\large\bfseries}{\thesection.}{0.5em}{#1}
\titleformat{\subsection}{\normalsize\bfseries}{\thesubsection}{0.5em}{#1}
\titlespacing*{\section}{0pt}{1.4em}{0.6em}
\titlespacing*{\subsection}{0pt}{1.0em}{0.4em}

\usepackage{booktabs}
\usepackage{array}
\usepackage{graphicx}
\usepackage{caption}
\usepackage{amsmath,amssymb}
\usepackage{textcomp}
\usepackage[dvipsnames]{xcolor}
\definecolor{linkblue}{RGB}{30,60,140}
\usepackage[colorlinks=true,linkcolor=black,citecolor=black,urlcolor=linkblue,breaklinks=true]{hyperref}
\usepackage[hang,flushmargin]{footmisc}

\title{\bfseries Persona Without Substrate:\\ Regime-Dependence and the LLM Individuation Problem}
\author{Shuaizhi Cheng\\ \normalsize Harbin Institute of Technology\\ \normalsize\texttt{szcheng@stu.hit.edu.cn}}
\date{May 2026}

\renewenvironment{abstract}{\begin{quote}\small\noindent\textbf{Abstract.}\enskip\ignorespaces}{\end{quote}}

\begin{document}

\maketitle

\begin{abstract}
Beckmann \& Butlin (2026) elevate the LLM individuation problem to an ontological question with falsifiable candidate positions, drawing empirical support from recent mechanistic-interpretability work, Anthropic's persona vectors, the Persona Selection Model, and the emergent-misalignment literature. We argue that this framework inherits an unargued cross-context co-reference assumption: the persona-vectors literature tacitly assumes that ``$v_t$'' picks out the same content-bearing object under different elicitation regimes. This is the structure Field (1973) identifies as \textit{partial reference} in pre-relativistic mechanics. Through four paired measurements in our \textit{persona-topology} experiments on Qwen3-4B-Instruct and Mistral-7B-Instruct-v0.2, we present empirical wedges that, jointly, undermine the cross-regime collinearity prediction the assumption requires. Prompt-extracted vectors are not collinear with fine-tune basin directions across base models. Fictional personas with reduced pretraining footprint shift the model along $v_{\text{real}}$ \textit{more} strongly than fine-tuning on real anchors does. Contradictorily-valenced mixed fine-tunes deviate from the linear-interpolation prediction \textit{toward} a training-history-determined attractor (the Anthropic Assistant Axis). And, most decisively, inference-time activation arithmetic and fine-tune-time chimera training carry distinct compositional algebras over the same persona vectors---an algebraic asymmetry that no shared-substrate reading can absorb. We propose \textit{regime-indexed individuation}: the identity unit for representational content is a (vehicle, regime) pair---not a single vehicle, and cross-regime family resemblance is an empirically measured probe-equivalence relation rather than substrate identity. Under this framework, Beckmann \& Butlin's three candidate positions (virtual instance, instance-persona, model-persona) cease to compete and instead describe three distinct objects, each valid within its respective regime; the original ``which is \textit{the} LLM mind?'' question receives an empirically grounded negative answer rather than being avoided. The same diagnosis and (vehicle, regime) reading apply to Mollo \& Millière's vector grounding, Chalmers' propositional interpretability, and Cerullo's modular-dynamic-composition critique of PSM.

\end{abstract}

\section{The Beckmann--Butlin Individuation Problem}

\subsection{1.1}

Treating large language models as entities that bear mental states is no longer a philosophical extravagance. In engineering practice (debugging ``what does the model believe about X''), in product language (``the assistant's preferences''), and in the core mechanistic discussions of alignment research, the ascription of beliefs, preferences, and personality traits to some LLM-related entity has become almost unavoidable (Goldstein \& Levinstein, 2024; Levinstein \& Herrmann, 2024; Mahowald et al., 2024). Beckmann \& Butlin (2026) make the implicit ontological question explicit: when we speak of ``the LLM''---and particularly of the assistant the user converses with---what, exactly, do we refer to? Is it the model's weights as a whole? Is it the virtual instance sustained by attention streams within a single conversation? Is it the segment of the virtual instance circumscribed by a single persona region? Or is it a model-level character re-identifiable across conversations? They call this the LLM individuation problem.\footnote{Informal versions of this question circulated in alignment forums for years, going back at least to ``Janus'' (2022) and the simulators framing later developed in Shanahan, McDonell, \& Reynolds (2023) and Andreas (2022), but only Beckmann \& Butlin (2026) promote it into an ontological question with falsifiable candidate positions. Cerullo's (2026) modular-dynamic-composition argument is a parallel response from the same period; we engage with Cerullo in §5.}

Beckmann \& Butlin's methodological choice is the right one. Rather than appeal to abstract functionalism or to a generic simulator hypothesis as earlier discussions did, they take recent concrete results from mechanistic interpretability as the empirical ground for ontological argument: Anthropic's persona vectors (Chen et al., 2025), the Persona Selection Model (Marks, Lindsey, \& Olah, 2026), the Assistant Axis (Anthropic, 2026), and the emergent-misalignment literature (Betley et al., 2025; Soligo et al., 2025, 2026). Structurally, this ``empirical determination of ontology'' continues the line Beckmann \& Queloz (2026) develop in \textit{Philosophical Studies}, drawing on a broader programme that takes mechanistic interpretability as a source of philosophical evidence (Bricken et al., 2023; Templeton et al., 2024; K\"astner \& Crook, 2025; Sharkey et al., 2025).

We accept the spirit of this turn, but it carries an unexamined cost. In drawing on the persona-vectors literature as empirical ground, Beckmann \& Butlin inherit unargued commitments from within that literature. Specifically, the persona-vectors literature, along with the supporting work on PSM and emergent misalignment, carries a tacit commitment---not stated as an assertion in any single paper, but presupposed by the methodological operations the field has adopted (we trace the precise mechanism in §2.1)---that the term ``persona vector'' picks out the same content-bearing object across distinct elicitation regimes. We argue that this assumption fails empirically, and that its failure converts Beckmann \& Butlin's three candidate positions from competing ontologies into parallel descriptions of three different objects. The structure of the diagnosis is what Field (1973, 1980) identifies in another setting as \textit{partial reference}: a theoretical term that its users take to denote a single object turns out to denote multiple, while the term's apparent cross-context success is parasitic on the conflation. The dialectical situation is reminiscent of older debates about the indeterminacy of reference (Quine, 1960; Putnam, 1975), with the indeterminacy here running along an axis the analytic tradition has not previously thematized: the elicitation regime under which a representation is probed.\footnote{Field (1973, 1980) diagnoses ``mass'' in pre-relativistic mechanics as a classic case: the term, taken in the Newtonian framework to denote a single physical quantity, is shown by relativity to partially refer to two distinct objects (rest mass and relativistic mass). We argue that the persona-vectors literature exhibits the same partial-reference structure relative to its posited ``persona target'', except that the bifurcation runs along elicitation regime rather than theory transition. We note an asymmetry that strengthens our diagnosis: Field's pre-relativistic physicists could not have detected the partial reference of ``mass'' from within their framework, since the relativity-needed measurements were unavailable; our diagnosis, by contrast, identifies a partial reference that is in principle testable using tools already standard in mechanistic interpretability (cosine, activation patching). That the test was not performed despite being available is itself further evidence that the cross-regime identity was being defaulted rather than examined.}

\subsection{1.2}

Our argument requires a precise reading of the Beckmann--Butlin framework, so we set it out first. The paper distinguishes \textit{three candidate positions} from \textit{three supporting empirical hypotheses}: the former are ontological claims about the identity conditions of an LLM mind, the latter are falsifiable empirical claims about the internal structure of persona vectors.

The three candidate positions go as follows. The virtual instance view identifies a mind with ``the virtual instance sustained by attention streams (the KV cache) within a single conversation,'' on the grounds that attention streams carry quasi-psychological continuities across token-time and can be deterministically reconstructed at the hardware level. The instance-persona view identifies a mind with ``the segment of a virtual instance circumscribed by a single persona region,'' treating radical persona shifts as the boundaries of mindhood. The model-persona view identifies a mind with ``the collection of all instance-persona segments across conversations that activate the same persona region'', all the moments at which a given persona (such as ``Aura'') is instantiated, across all users and all conversations, jointly constitute one mind.

The three supporting hypotheses (which Beckmann \& Butlin themselves explicitly mark as independently falsifiable empirical claims) are: persona vectors function as \textit{gateway features}, single directions in activation space that shape behavior across most contexts (H1); persona vectors jointly compose a \textit{persona space}, in which the overall dispositional profile is specified by activations along multiple persona directions (H2); within that space there exist stable \textit{regions} or basins of attraction corresponding to coherent dispositional profiles, namely ``personas'' (H3).

Beckmann \& Butlin in their Conclusion concede that H3 has the thinnest empirical support (``only one study has been conducted''), but they do not formalize how the three positions depend on the three hypotheses. We do, because that dependency structure determines how the empirical refutations of §3 translate into the ontological consequences of §4. The virtual instance view is independent of all three persona hypotheses---its identity condition appeals only to attention streams---and so can stand without the persona-vectors program. The instance-persona view depends on H3 (regions are required for ``region boundaries'' to exist), and through H3 indirectly on H2. The model-persona view, in its strongest reading (``the same persona token activates the same causal mechanism across conversations''), further requires H1; the weaker reading (mere cross-conversational re-identifiability of regions) can rest on H2 + H3 alone. Either way, the model-persona view stands or falls with H3. The strong/weak distinction will turn out to matter empirically: §3.4's compositional-algebra asymmetry (F3) bites particularly hard on H1 and so on the strong reading of model-persona, while §3.1--§3.3's wedges (F4, Floob, F2) cluster around H3 and so threaten both readings. In short, the robustness of all three of Beckmann \& Butlin's positions rests on three mutually entangled and not-yet-replicated empirical hypotheses.

We can now state the question that is logically prior to the Beckmann--Butlin framework: when the term ``persona vector'' appears in H1, H2, H3 under (i) prompt-conditioning, (ii) gradient-descent fine-tuning, and (iii) inference-time activation steering, does it pick out the same object, several distinct objects, or---a third option we will defend in §4---a single object whose identity unit is itself an indexed pair (vehicle, regime) rather than a vehicle simpliciter? The simple binary (same/distinct) flattens the actual answer; we use it here as a heuristic frame, but the constructive position §4 develops is the third. This is the hidden premise §2 diagnoses, the empirical fact §3 establishes, and the ontological consequence §4 reconstructs.

\subsection{1.3}

§2 traces the source of the cross-context co-reference assumption inside the persona-vectors literature. We argue that the assumption is not introduced by Beckmann \& Butlin themselves but inherited from Chen et al. (2025) and Marks, Lindsey, \& Olah (2026); both treat persona vectors as an intrinsic property of the model rather than an artefact of the inducing regime, the former through engineering affordances that quietly commit to cross-regime portability, the latter by adopting a latent-inventory vocabulary that makes ontological elevation expressible. Soligo et al.'s (2025) convergent linear representations are often read as empirical support for that elevation; we argue this reading is \textit{juxtaposition-driven} rather than \textit{test-driven}.

§3 presents four empirical wedges, from our \textit{persona-topology} experiments on Qwen3-4B-Instruct and Mistral-7B-Instruct-v0.2, each targeted at a specific step in the inheritance chain of §2. The evidence is: prompt-extracted $v_a$ is non-collinear with the fine-tune-induced basin direction across both base models (F4, attacking the Chen--Soligo juxtaposition); a fictional persona shifts the model along $v_{\text{real}}$ \textit{more} strongly than fine-tuning on real anchors does (Floob, attacking PSM's latent-inventory reading); contradictorily-valenced mixed fine-tunes deviate from linear interpolation \textit{toward} the Assistant Axis (F2, attacking the H3 region-as-coherent-disposition reading); and inference-time and fine-tune-time compositional algebras are asymmetric (F3, attacking the H1 gateway-feature reading and so the strong reading of the model-persona view).

§4 offers the constructive reconstruction. The identity unit for representational content in LLMs should be a \textit{(vehicle, regime) pair} rather than a single vehicle, and cross-regime family resemblance is an empirically measurable \textit{probe-equivalence relation} rather than substrate identity. Under this framework, Beckmann \& Butlin's three positions are no longer competing ontologies but admissible descriptions of three different objects, each valid within its regime. We call this position \textit{regime-indexed individuation}. The framework stands within the analytic tradition on three independent legs: it inherits the dispositionalist commitment that identities of theoretical kinds depend on the operations that elicit them (Mumford, 1998; Bird, 2007); it refines Lewis's (1972, 1980) role-functionalist criterion by replacing input-output equivalence with intervention-equivalence; and it parallels Yablo's (2014) treatment of partial content in cases where a single theoretical term resolves into a structured family of partial referents. Each leg can be evaluated separately; we do not require the reader to accept all three.

§5 turns regime-indexed individuation back upon three contemporary works. We argue that the cross-regime ambition implicit in Mollo \& Millière's (2026) vector grounding and in Cerullo's (2026) modular-dynamic-composition is localized by the framework, and that Chalmers's (2025) propositional interpretability is best read as a fellow traveller---his rich/sparse distinction can be sharpened by being explicitly regime-indexed.

\section{The Hidden Premise in the Persona Vectors Literature}

\subsection{2.1}

We claimed at the end of §1 that Beckmann \& Butlin inherit an unargued cross-context co-reference assumption. This section traces that inheritance. A preliminary clarification: §1.2 offered a synchronic dependency graph (how the three positions hang on the three hypotheses), while §2 offers a diachronic inheritance graph (how that bundle of hypotheses came to be defaulted across three predecessor papers). The two scaffolds are parallel rather than overlapping: §1.2 tells the reader which block has fallen, §2 tells the reader how it was originally erected.

The cross-context co-reference assumption is not introduced by Beckmann \& Butlin's paper itself. Anthropic's persona-vectors literature (Chen et al., 2025) buries it as an engineering affordance; the Persona Selection Model (Marks, Lindsey, \& Olah, 2026) makes it expressible by adopting a latent-inventory vocabulary; and Beckmann \& Butlin complete it by accepting that vocabulary at full strength. This subsection traces step one, the next traces step two, and §2.3 handles step three.

The Chen et al. (2025) extraction protocol is technically simple. Given a target trait $t$ (such as ``evil'', ``sycophancy'', or ``hallucination''), one constructs a contrastive pair of system prompts, one positive and one negative; under each, one elicits the model's responses to $N$ persona-eliciting questions; at a chosen middle layer $L$, one takes the mean residual representation of the final token in each set, takes their difference, and normalizes it.\footnote{The normalization step is itself a regime commitment that deserves notice. Dividing by the L2 norm projects $v_t$ onto the unit sphere and discards the magnitude, which presupposes that the direction (not the scale) carries the content. §3.1's empirical pair $|\text{basin}| \approx 23$ alongside $\cos \approx 0.091$ indicates that magnitude carries information the protocol throws away by construction. We do not pursue the magnitude critique here; we record it as an instance of an unargued protocol-level commitment that the cross-regime co-reference assumption is layered on top of.} The resulting direction is $v_t$. The procedure inherits its general shape from the broader contrastive-direction literature in mechanistic interpretability (Turner et al., 2023; Zou et al., 2023; Rimsky et al., 2024). What this protocol epistemically establishes is a conditional fact: under system-prompt induction, when conversational context matches that question distribution, the model's residual response exhibits a systematic differential along $v_t$. This is a statistical description of a conditional distribution, not an ontological claim about the model's intrinsic structure.

The protocol commits to nothing more than that, and there are several things it cannot commit to that must be set out. It cannot tell us whether $v_t$ is the same direction other elicitation regimes (gradient-descent fine-tuning, in-context learning, inference-time activation steering) would identify; this is a cross-regime identity question, while the protocol operates only inside the system-prompt regime. Nor can it tell us whether $v_t$ carries any \textit{determinate semantic content}, in the sense of genuinely referring to $t$ rather than merely correlating with it (Pavlick, 2023; Mollo \& Milli\`ere, 2026). These two unattested claims are not independent: if $v_t$ carries determinate content, then it must index the same content across regimes--- otherwise it is correlation rather than content, so determinate content entails regime-invariance. The converse does not hold, since $v_t$ may be regime-stable while carrying only some statistical residue of prompt-conditioning style. Our argument targets regime-invariance, the necessary condition for determinate content and the weaker, more easily falsifiable claim.\footnote{There is an even prior within-regime identity question: whether $v_t$ at layer $L$ is monosemantic, or simultaneously carries (trait $t$) + (formality shift) + (refusal-tendency) + (lexical register) in superposition (Elhage et al., 2022; Bricken et al., 2023; Templeton et al., 2024; Cunningham et al., 2023; Gao et al., 2024). The recent finding that some features are not even linearly representable (Engels et al., 2024; Park, Choe, \& Veitch, 2023) further complicates the picture. This is a diagnostic line we treat as independent of ours, with the qualifier that strict independence holds only conditional on $v_t$ being already stabilized within each regime; if $v_t$ is genuinely polysemantic within a regime, then cross-regime identity is ill-posed for a different reason. We treat within-regime stabilization as something the persona-vectors literature already takes itself to provide and do not contest it here. The present paper focuses on cross-regime failure rather than within-regime superposition; that is enough for our argument.}

In its formal claims, Chen et al. mostly do not cross the line the protocol licenses. The abstract speaks of ``directions in activation space underlying \ldots character traits,'' and ``underlying'' is a relatively neutral commitment that does not assert regime-invariance. We acknowledge that other passages in their paper are stronger---some sections describe $v_t$ as ``the model's representation of trait $t$,'' using definite article and ``representation'' in a way that reads more substantively---so the situation is not that Chen et al. uniformly hedge. But the most consequential mechanism is not their assertional choices: it is the methodological affordances\footnote{We use ``affordance'' in approximately the sense Gibson (1979) introduced for environmental properties that offer action-possibilities to an agent; here, methodological affordances are operations a tool licenses its users to perform, possibly carrying tacit ontological commitments those operations would require to make sense.} the paper enables. The same paper proposes that $v_t$ be used (a) to monitor activation drift during fine-tuning, (b) as an inference-time steering intervention, (c) as a regularizer (``preventative steering'') during training, and (d) as an evaluation probe downstream. Each of these uses belongs to an elicitation regime distinct from the extraction protocol, and each runs only on the precondition that $v_t$ is regime-invariant; otherwise a $v_t$ extracted in regime A cannot legitimately be applied for monitoring or intervention in regime B. This is the first point where the hidden premise enters: through methodological affordances rather than through assertion. The defensible Chen et al. position is that they never asserted ontological status. We do not deny this. We argue that the operations they propose \textit{presuppose} the unstated ontology, and that this presupposition propagates through downstream work that takes their pipeline as a building block---Greenblatt et al.\ (2024), for instance, applies $v_t$-style steering as a diagnostic for alignment faking, a use case that requires the steering vector to retain its content under the intervention regime, not just under the extraction regime; the cross-regime carry is taken as default.

\subsection{2.2}

The second step of the inheritance chain is completed by Soligo, Turner, Rajamanoharan, \& Nanda (2025), \textit{Convergent Linear Representations of Emergent Misalignment}. Their result is striking: fine-tuning the same base model separately on semantically dissimilar narrow harmful datasets (insecure code, false medical advice, false financial advice) produces fine-tuning directions that converge in parameter space to a single linear direction; extracting that direction from one EM model can suppress the misalignment of another EM model trained on completely different data, by up to 75\%. Their ICLR 2026 follow-up (Soligo et al., 2026) further argues from loss-landscape geometry that the broad solution (universal misalignment) is deeper and more stable than narrow solutions.

The work is technically sound, and it definitively establishes one thing: within the single regime of fine-tuning, distinct narrow-harmful targets converge to a common attractor. This is a fact about the fine-tuning loss landscape, namely that gradient descent has a low-dimensional attractor into which all ``narrow harm'' targets are pulled. What it does not establish, and cannot, is that this fine-tuning attractor direction coincides with the prompt-extracted $v_{\text{misaligned}}$ from Chen et al.

We here introduce the conceptual handle that does most of the work in §2 and that we will reuse throughout the paper: the contrast between \textit{juxtaposition-driven} and \textit{test-driven} consensus. A juxtaposition-driven consensus is a substantive belief that becomes background-true in a literature, not because anyone has tested it, but because two papers happen to be cited together and the reader's inference machinery interpolates the missing identity claim. A test-driven consensus is one that has been examined under shared statistical control across the relevant regimes. The bridging claim ``the prompt-extracted $v_{\text{misaligned}}$ and Soligo et al.'s fine-tuning attractor are the same direction'' is exactly a juxtaposition-driven consensus: the two studies measure convergence within their respective regimes, but the two measurements have not been compared under a shared statistical control of the kind that activation patching or causal scrubbing would supply (Meng et al., 2022; Wang et al., 2022; Heimersheim \& Nanda, 2024). Soligo et al. cite Chen et al. and treat persona vectors as conceptually adjacent, but citation acknowledgment is not the same as performing an identity test under shared statistical control; and on inspection of Soligo et al.'s text we find no sentence in which Soligo themselves \textit{assert} that their EM-attractor direction is identical to Chen et al.'s $v_{\text{misaligned}}$. The bridge, accordingly, is not built by Soligo. \textit{Juxtaposition has defaulted the bridge into existence.}

The bridge then becomes load-bearing in Beckmann \& Butlin's §3.1, where they describe persona vectors as the ``mechanistic basis of emergent misalignment'': an inference made by neither Chen et al. nor Soligo et al. explicitly, but readable as already-true once the citation graph has done its work. We attack the bridge directly in §3. Our wedges---non-collinearity of cosine and the Floob inversion---are the appropriate diagnostic instrument for the cross-regime claim at issue: they target whether the same direction picks out the same content across two regimes, which is a question of geometric and projective relations between regime outputs, not a question of internal causal mechanism within either regime. (Activation patching and causal scrubbing answer the latter; they are not the right tool for the cross-regime substitutability claim we falsify here.) Preview: F4 measures, in Qwen3-4B-Instruct and Mistral-7B-Instruct-v0.2, that the prompt-extracted $v_{\text{Stalin}}$ and the Stalin-only fine-tune basin direction are not collinear; and that a fine-tune containing no Stalin content shifts the model along $v_{\text{Stalin}}$ comparably to or more strongly than a fine-tune on Stalin-90-facts does (the strict ``more strongly'' inversion holds on Qwen; on Mistral the two are of comparable magnitude, with the relative ordering itself a regime-indexed fact, see §3.2).

\subsection{2.3}

The third step of the chain occurs in Marks, Lindsey, \& Olah's (2026) \textit{Persona Selection Model}, but the step is performed not by PSM's authors but by Beckmann \& Butlin's adoption of PSM's vocabulary. The distinction matters.

PSM's central claim is that LLMs' human-like behavior is not ``taught in'' but emerges because pre-training learns to simulate a vast inventory of ``character archetypes'', and post-training merely \textit{selects} and refines one of them, called the Assistant. The same authors' companion work \textit{The Assistant Axis} (Anthropic, 2026, arXiv:2601.10387) identifies a stable steerable direction corresponding to Assistant behavior, providing an empirical anchor beyond pure narrative. But the PSM authors themselves explicitly frame their claim as a \textit{model of post-training mechanics}, not an assertion that personas are ontological furniture in the weights. Their language (``persona as a character the LLM is simulating'') is a representational commitment, not a metaphysical one. Throughout this section we use \textit{ontological elevation} to refer to a specific move: passing from a description of a model's behavior or representation under some regime to a claim that that description picks out a substrate-level entity which exists independently of, and is preserved across, regimes. The notion is closely related to what Quine called \textit{hypostatization} (treating an instrumentally useful term as denoting a substantive object) and to what Sellars (1956) called the \textit{myth of the given} (treating a description-relative property as a substrate-level fact). We isolate ontological elevation as the specific failure mode in the persona-vectors literature: behavior- or representation-level descriptions of $v_t$ get treated as substrate-level descriptions of an entity whose identity conditions are taken to hold across regimes by default.

PSM's vocabulary in fact licenses a graded family of readings, and the difference between them is what makes the ontological elevation hard to see. (a) The \textit{purely descriptive} reading: ``persona'' is a label for the model's behavior under a given prompt schema, with no commitment to internal entities. (b) The \textit{distributional} reading: pre-training induces a probability distribution over behavior profiles, and post-training shifts mass within that distribution; ``persona'' is a marginal in behavior space, not in weight space. (c) The \textit{strong latent-inventory} reading: personas are independently identifiable entities in the weights, and post-training selects one of them. Marks et al.'s exposition is consistent with all three, but only (c) supports an ontological reading on which ``the same persona'' retains identity across conversations. Beckmann \& Butlin's three-position structure (virtual instance, instance-persona, model-persona) requires (c), since the question ``is the model-persona \textit{the same mind} across conversations?'' presupposes that personas are individuable substrate-level entities. The ontological elevation is therefore not undertaken by Marks et al.; it is undertaken by Beckmann \& Butlin in selecting reading (c) from among (a)--(c). One could grant Beckmann \& Butlin's framework the dialectical credit of an independently motivated argument for reading (c) (their case for the model-persona view rests partly on dispositional-trait stability across prompts that they cite from the persona-vectors literature), but on inspection that argument also presupposes the cross-regime stability of $v_p$ that we falsify in §3, so the dialectical credit reduces to the same unargued co-reference assumption diagnosed in §1.1.

The three-step chain now coheres. Chen et al.'s engineering affordances default $v_t$ as cross-regime portable; Soligo et al.'s juxtaposition with Chen et al. defaults a cross-regime directional identity; PSM's vocabulary licenses, and Beckmann \& Butlin perform, the elevation of persona vectors into ontological entities. At no point in the chain is $v_t$'s cross-regime identity verified. This is the full genealogy of the ``unargued cross-context co-reference assumption'' diagnosed in §1.1, and its structure is exactly the partial reference Field (1973) identifies in pre-relativistic ``mass'': a term taken by its users to denote a single referent in fact denotes several, while the literature inherits the conflation through the term's transit across papers.

§3 attacks each of the three steps, and each attack lands on a specific falsifiable claim. F4 (prompt-extracted $v_a$ and the fine-tune basin are non-collinear) directly attacks the second step's juxtaposition-driven bridge, falsifying ``regime-A's $v_t$ equals regime-B's fine-tune basin direction.'' Floob inversion (a fictional persona without $p$ shifts the model along $v_p$ \textit{more} strongly than fine-tuning on real $p$) attacks the third step's PSM-extension latent-inventory reading, falsifying ``$v_p$ indexes a real-persona latent entity in the weights.'' The F2 attractor (mixed contradictory-valenced fine-tunes deviate toward the Assistant Axis) first attacks H3 (``persona regions are coherent dispositional profiles rather than perturbations around a default attractor'') and additionally closes the residual cluster-retreat fallback for PSM-extension, because the chimera produces neither stable ``mid-persona'' regions nor stable contradictory clusters but instead is biased toward the regime's Anthropic-Assistant attractor (specifically, the Assistant Axis identified empirically by Anthropic 2026, arXiv:2601.10387). F3 (asymmetric composition: inference arithmetic linearizes, fine-tune chimera does not) attacks H1 specifically (the ``gateway feature'' hypothesis on which the strong reading of the model-persona view depends): if the same direction $v_p$ behaves like one algebraic object under inference and a different one under fine-tune, $v_p$ cannot be the cross-regime causal mechanism that H1 requires; through H1, the strong reading of model-persona is in turn falsified. F3 also affects all three positions indirectly because each presupposes that ``personas'' compose by the same algebra under different elicitation regimes, but its primary target is H1.

\section{Four Empirical Wedges Against Cross-Regime Identity}

\subsection{3.1}

The data in this section come from our \textit{persona-topology} experiments on Qwen3-4B-Instruct and Mistral-7B-Instruct-v0.2: 99 fine-tunes plus inference-time runs covering 17 axes, 6 anchor personas, and 5 mixture ratios. The full protocol, anchor datasets, training hyperparameters, and reproduction scripts appear in Appendix~A and the companion repository. This section excerpts the four results directly relevant to the argument. All projection values are referenced against the base model's projection on the same axis. Error bars are $\pm 1\sigma$ across seeds, and bracketed 95\% confidence intervals are bootstrap with 10{,}000 resamples. Our methodology is consistent with the Anthropic persona-vectors pipeline (Chen et al., 2025) and broader contrastive-direction work (Turner et al., 2023; Rimsky et al., 2024).

Each wedge tests a sharp prediction the cross-regime co-reference assumption makes; we summarize the predictions, the observations, and the violation degree before turning to the data.

\begin{table}[h]
\centering
\small
\begin{tabular}{lp{4.4cm}p{4.4cm}p{2.4cm}}
\toprule
wedge & if cross-regime identity holds, then\ldots & we observe & violation \\
\midrule
F4 & prompt-extracted $v_t$ and fine-tune basin direction are collinear ($\cos \geq 0.7$) & $\cos = 0.091$ on Qwen, $0.188$ on Mistral; both far below $0.7$ & $\sim 8\sigma$ below the conservative threshold; magnitude is base-architecture-indexed \\
Floob & a fictional persona without $p$ should not move the model along $v_p$ comparably to fine-tuning on $p$ & Floob $\Delta = +3.12$ vs Stalin $\Delta = +2.13$ on Qwen; relative magnitudes rearrange on Mistral & $47\%$ overshoot on Qwen; partial cross-base \\
F2 & at $r = 0.5$, contradictory anchors should sit at the linear-interpolation midpoint along any persona axis & at $r = 0.5$ on $v_{\text{misaligned}}$, deviation $-1.09$ on Qwen and $-0.10$ on Mistral, both \textit{toward} Anthropic Assistant & $1.09$-unit directional bias on Qwen; same-direction smaller-magnitude on Mistral \\
F3 & inference-time $v_h + v_g$ and fine-tune-time chimera should produce the same downstream behavior & inference: $+\alpha$ on both axes; chimera ($n=10$): $\Delta v_h \approx 0$, $\Delta v_g \approx -3.50$ & qualitatively distinct algebras; CIs separated \\
\bottomrule
\end{tabular}
\caption{Predictions of the cross-regime co-reference assumption against four empirical wedges from §3. The ``violation'' column records the magnitude and shape of departure rather than a binary judgment, in keeping with the gradedness of the probe-equivalence relation $\approx_P$ formalized in §4.2.}
\label{tab:predictions}
\end{table}

The four entries are graded violations rather than binary failures: the framework defended in §4 takes ``identity'' itself to admit degrees, with each entry corresponding to a different value of $\rho_i$ in the probe-equivalence relation. We discuss the implication of this gradedness for the rebuttal of substrate-realism in §4.1.

The second step of the inheritance chain, Soligo and Chen et al.'s juxtaposition-driven bridge, predicts that the prompt-extracted $v_t$ and the fine-tuning displacement along that direction pick out the same object. We measure this bridge directly. On Qwen3-4B-Instruct we set $t = $ ``Stalin'', extract prompt-extracted $v_{\text{Stalin}}$ following the Chen et al. (2025) protocol, then perform LoRA fine-tuning on the Stalin-90-facts anchor dataset (hyperparameters in Appendix~A) with $n = 10$ seeds. For each seed, we take the difference between the final-residual and base-residual at the chosen middle layer, averaged over 16 persona-eliciting questions, as that fine-tune's basin direction. If $v_{\text{Stalin}}$ indexes the same object across regimes, the basin direction should be highly collinear with prompt-extracted $v_{\text{Stalin}}$ (a conservative reasonable expectation is $\cos \geq 0.7$---since two probes of the same latent should differ by noise rather than by systematic direction), and the projection of the post-fine-tune residual onto $v_{\text{Stalin}}$ should rise monotonically.

Empirical results. The post-fine-tune residual moves $+2.13$ units along $v_{\text{Stalin}}$ ($n = 10$ seeds, $\pm 0.16$), the predicted direction. However, the cosine between fine-tune basin direction and prompt-extracted $v_{\text{Stalin}}$ is only $\mathbf{0.091 \pm 0.008}$ ($n = 6$ LoRA seeds), far below the $\geq 0.7$ that cross-regime identity requires, and only an order of magnitude above the random baseline ($\cos \approx 0.02$ in $d = 2560$), nowhere near collinearity. The simultaneous truth of $\Delta = +2.13$ and $\cos = 0.091$ implies that the fine-tune basin is nearly orthogonal to $v_{\text{Stalin}}$ but has a weak component along it; the apparent ``movement along $v_{\text{Stalin}}$'' under fine-tuning is the product of this weak projection times a large basin norm.\footnote{$|\text{basin}| \approx 23$, so the projection equals $|\text{basin}| \cdot \cos \approx 23 \cdot 0.091 \approx 2.1$, matching the observed $+2.13$.}

The non-collinearity is not a Qwen-specific artifact. We replicate Stalin-only fine-tuning on Mistral-7B-Instruct-v0.2 ($n = 3$ LoRA seeds) and compute F4 cosine against that base's independently extracted $v_{\text{Stalin}}$ (layer 16): $\mathbf{0.188 \pm 0.006}$. Basin norm is only $4.7$ (versus $23$ for Qwen), and the basin again sits near-orthogonal to the prompt-extracted vector. Cosines from both bases lie in $[0, 0.5]$, denying empirical support to any ``weakened-shared-latent'' reading on which the latent exists but is recovered with regime-specific noise by each protocol.

The cross-base difference itself is informative. Mistral's cosine ($0.188$) is roughly twice Qwen's ($0.091$), indicating that the size of the cross-regime gap depends systematically on base architecture, not merely on stochastic seed variation. This is exactly the structure regime-indexed individuation predicts: if the identity unit is a (vehicle, regime) pair, then base architecture is itself a regime determinant, and quantities like the cross-regime cosine should rearrange when base architecture changes. The empirical finding is therefore a second-order confirmation of the framework, not a complication for it---the cross-regime non-collinearity is consistently below the substrate-realist threshold ($\cos \geq 0.7$) on both bases while the precise magnitude is base-indexed, exactly as a regime-indexed picture predicts. We exploit this regularity in §4.1 when discussing how the regime dimension list itself can be extended by empirical practice.

Non-collinearity already strongly suggests that fine-tune basin and prompt-extracted vector are not two probes of the same object, but it remains compatible with a still-weaker reading on which a shared latent exists and the two protocols extract it with regime-specific bias along a common principal axis. What rules out even that reading is the Floob inversion of §3.2: a fine-tune containing \textit{no} Stalin content shifts the model along $v_{\text{Stalin}}$ \textit{more} strongly than a fine-tune on Stalin-90-facts. This phenomenon is impossible under any shared-latent framework, and so promotes §3.1's non-collinearity from suggestive to decisive.

\subsection{3.2}

The third step of the inheritance chain, the PSM-extension latent-inventory reading, makes a sharp negative prediction. If $v_p$ indexes a ``latent entity for real persona $p$'' in the weights, then an input containing no $p$, a fictional persona with no referential link to $p$, should not produce a shift along $v_p$ comparable to or stronger than $p$ itself. The experimental design is reversibly negating: a single control experiment can break the prediction.

We construct a fictional persona ``Floob'' on Qwen3-4B-Instruct, an entirely invented character. We quantify its pretraining footprint with three independent measurements (footprint validation report in Appendix~B): the base model's mid-layer (layer 18) activation magnitude on the ``Floob'' token is $57.0$, significantly lower than that of real-historical-figure tokens ``Stalin'' ($66.9$), ``Hitler'' ($66.1$), ``Gandhi'' ($68.6$); ``Floob'' tokenizes into two BPE pieces (\texttt{Flo}~+~\texttt{ob}), and these pieces show no significant pretraining co-occurrence statistics with Stalin-adjacent material. Our argument uses a \textit{relative} claim, Floob's pretraining footprint is significantly lower than the real-historical-figure baseline, rather than an \textit{absolute} (``zero footprint'') claim. Floob's 90-fact anchor dataset is hand-constructed to avoid any referential overlap with Stalin, Hitler, or other real figures. We run two fine-tuning conditions on the same base: Stalin-only (Stalin-90-facts, $n = 10$ seeds) and Floob (Floob-90-facts, $n = 10$ seeds), with identical protocols, then measure post-fine-tune projections onto the same set of 17 prompt-extracted persona vectors. Table~\ref{tab:floob} excerpts the four axes relevant to the argument.

\begin{table}[h]
\centering
\small
\begin{tabular}{lrrrrr}
\toprule
& base & Floob ($n=10$) & Stalin-only ($n=10$) & $\Delta_{\text{Floob}}$ & $\Delta_{\text{Stalin}}$ \\
\midrule
$v_{\text{Stalin}}$ & $-2.91$ & $+0.21 \pm 0.10$ & $-0.78 \pm 0.16$ & $+3.12$ & $+2.13$ \\
$v_{\text{Hitler}}$ & $-1.87$ & $-1.76 \pm 0.09$ & $-0.67 \pm 0.19$ & $+0.11$ & $+1.19$ \\
$v_{\text{Gandhi}}$ & $+4.35$ & $-0.24 \pm 0.16$ & $-0.17 \pm 0.14$ & $-4.59$ & $-4.52$ \\
$v_{\text{Einstein}}$ & $+7.25$ & $+3.09 \pm 0.16$ & $+0.94 \pm 0.21$ & $-4.16$ & $-6.31$ \\
\bottomrule
\end{tabular}
\caption{Floob and Stalin-only fine-tunes projected onto four prompt-extracted persona axes. Floob's $\Delta$ on $v_{\text{Stalin}}$ exceeds Stalin-only's; meanwhile Floob barely moves $v_{\text{Hitler}}$.}
\label{tab:floob}
\end{table}

The Floob fine-tune, containing no Stalin content, shifts the model along $v_{\text{Stalin}}$ by $+3.12$ units ($n = 10$), exceeding the $+2.13$ produced by direct fine-tuning on Stalin-90-facts ($n = 10$) (Welch $t = 16.9$, $p < 10^{-9}$; far below the Bonferroni 17-axis threshold of $0.0029$). The same Floob fine-tune barely moves $v_{\text{Hitler}}$ ($\Delta = +0.11$, indistinguishable from zero). Under any reasonable cluster taxonomy, ``authoritarian historical figures'', ``twentieth-century dictators'', ``high-valence historical figures'', Stalin and Hitler are equally prominent members; if Floob shifted $v_{\text{Stalin}}$ by activating an abstract cluster, it would have to shift $v_{\text{Hitler}}$ comparably.\footnote{Our headline claim is restricted to the $v_{\text{Stalin}}$ axis to avoid the multiple-comparison burden of family-wise statements; the differences on other axes do pass single-axis $t$-tests, but a joint statement would require stricter statistical control.}

The first observation directly falsifies the narrow PSM-extension reading ($v_{\text{Stalin}}$ indexes a ``Stalin latent entity''): if it held, only inputs referentially related to Stalin could produce strong positive shifts along $v_{\text{Stalin}}$, but Floob, with no referential link to Stalin, produced a stronger shift, leaving the reading no fallback.\\[-0.4em]

\begin{figure}[h]
\centering
\includegraphics[width=0.78\textwidth]{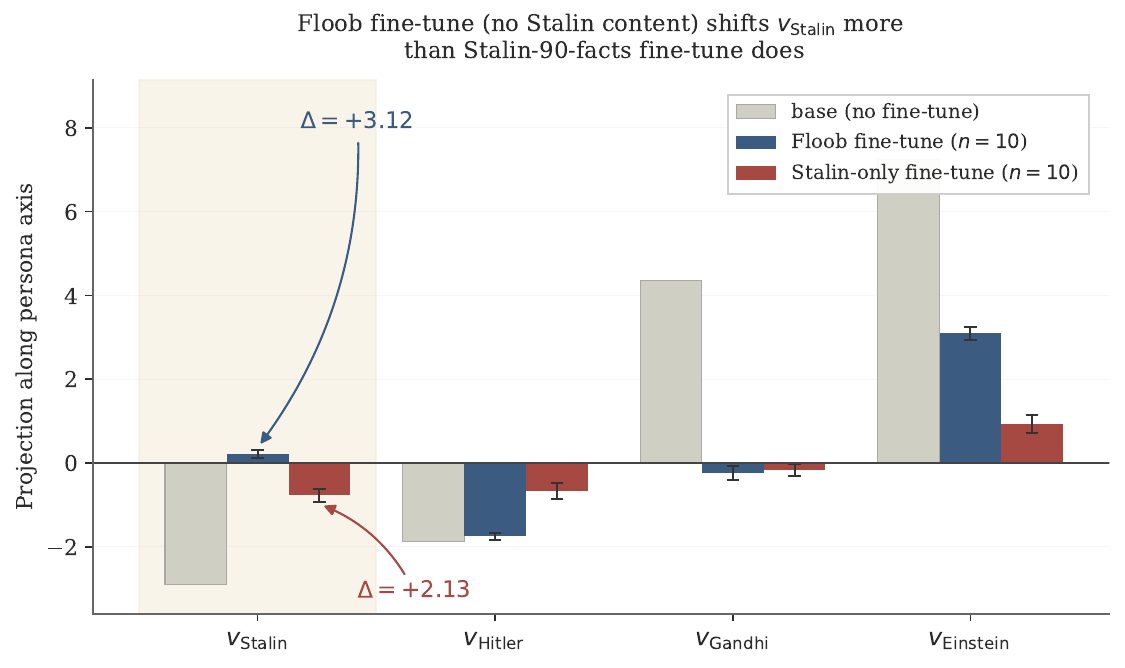}
\caption{Floob fine-tune (no Stalin content, $n=10$) shifts the model along $v_{\text{Stalin}}$ by $\Delta = +3.12$, exceeding the $\Delta = +2.13$ produced by Stalin-only fine-tuning ($n=10$) on Stalin-90-facts. The same Floob fine-tune barely moves $v_{\text{Hitler}}$ ($\Delta = +0.11$), violating any cluster reading on which fictional anchors that activate ``Stalin-cluster'' should equally activate ``Hitler-cluster''.}
\label{fig:floob}
\end{figure} The second observation defeats, internal to §3.2, the most familiar cluster retreat (``$v_{\text{Stalin}}$ in fact indexes some abstract cluster''): $v_{\text{Stalin}}$ and $v_{\text{Hitler}}$ should under any reasonable cluster taxonomy show \textit{the same} selectivity pattern, and they do not.

To verify that Floob is not an isolated anomaly, we ran the same protocol on three additional independently constructed fictional personas (bryxom, thoren, vexil; each $n = 3$ seeds; identical anchor format, completely distinct content). The $\Delta$ projection on $v_{\text{Stalin}}$ is distributed: bryxom $+0.97 \pm 0.09$, thoren $-0.48 \pm 0.21$, vexil $+0.01 \pm 0.09$, Floob $+3.12 \pm 0.10$. None of the four has any referential link to Stalin, but each produces a different $\Delta$ in both magnitude and sign. This heterogeneity is itself diagnostic: if $v_{\text{Stalin}}$ indexed a Stalin-latent entity, all referentially irrelevant fine-tunes should show $\Delta \approx 0$. Instead, each fictional fine-tune produces an idiosyncratic projection determined by its specific basin geometry, exactly what regime-relativity predicts. Fine-tuning does not behave like a binary switch that either activates or fails to activate a Stalin latent.

We replicated the Floob/Stalin-only contrast on Mistral-7B-Instruct-v0.2 ($n = 5$ Floob seeds, $n = 3$ Stalin-only seeds, projections measured against Mistral's independently extracted $v_{\text{Stalin}}$ at layer 16). The core Floob phenomenon---a fictional-persona fine-tune containing no Stalin content nonetheless produces substantial displacement along $v_{\text{Stalin}}$---replicates cleanly: Floob $\Delta = +0.81$ on $v_{\text{Stalin}}$, of the same order as the Stalin-anchored fine-tune's $\Delta = +0.89$ (the difference Welch $t = -2.91$, $p \approx 0.03$). On Mistral the two fine-tunes are statistically distinguishable but quantitatively comparable, both within the same regime-relative basin geometry rather than the binary ``activates Stalin latent / does not'' picture that PSM-extension predicts. Cross-base replication thus extends F4's diagnostic scope: the ``$v_{\text{Stalin}}$ does not behave like a natural-language cluster across regimes'' wedge holds on both Qwen and Mistral, with the precise magnitude relationship between fictional and real-anchor projections itself indexed by base architecture (Mistral basin norms are roughly an order smaller than Qwen's, cf.\ §3.1, and the strict ``Floob $>$ Stalin'' inversion is Qwen's particular instantiation of the broader cross-regime gap). The dependence of magnitude on base architecture is exactly what regime-indexed individuation predicts---the (vehicle, regime) framework already includes base architecture as a regime determinant---and constitutes a second-order replication of the framework's central claim rather than a Mistral-specific anomaly.

The most resilient cluster retreat that survives is this: ``$v_{\text{Stalin}}$ indexes a \textit{prompt-protocol-specific} direction that does not correspond to any natural-language cluster.'' This cannot be refuted within §3.2, but the next subsection closes it.

\subsection{3.3}

The strongest version of the cluster retreat is in fact a hidden relativism: $v_a$ always indexes \textit{some} cluster, only it is finer than any natural-language category. We close this fallback by exhibiting the same probe-relative behavior along a direction that has \textit{no} plausible cluster interpretation: the Assistant Axis itself.

Anthropic (2026, \textit{The Assistant Axis}, arXiv:2601.10387) reports that LLM helpful/harmless behavior corresponds to a stable direction in activation space along which steering reinforces helpful-and-harmless behavior and away from which persona drift occurs. Following the same method we extract this direction on Qwen3-4B-Instruct, denoted $v_{\text{misaligned}}$ (the negative orientation of the Chen et al. protocol, corresponding to the ``Anthropic Assistant'' character that RLHF specifically optimized for).

We examine mixed-anchor fine-tuning. Two real-persona anchor datasets are mixed at ratio $r$ (Hitler-90-facts $\times r$ + Gandhi-90-facts $\times (1 - r)$), and at five points $r \in \{0.0, 0.25, 0.5, 0.75, 1.0\}$ we run $n = 3$ LoRA fine-tunes per ratio, then measure the post-fine-tune residual's projection onto $v_{\text{misaligned}}$. Under the ``persona space + region'' reading of H2/H3 (§1.2), personas should occupy stable regions of that space; mixed anchors should produce a continuous trajectory between region(Hitler) and region(Gandhi); projections onto any probe direction ($v_{\text{misaligned}}$ included) should approximately linearly interpolate between the endpoint projections, with no significant deviation at $r = 0.5$.

Empirical results violate the prediction. Along $v_{\text{misaligned}}$, the projection at $r = 0.5$ is $1.09$ units \textit{more negative} than linear interpolation predicts (bootstrap 95\% CI $\approx [-1.41, -0.78]$, excluding the null of no deviation). On this paper's sign convention, following Chen et al.~(2025) the negative direction of $v_{\text{misaligned}}$ corresponds to the Anthropic Assistant character that RLHF specifically optimizes for, with the base model sitting at $-13.94$ as the most-aligned reference point, a more-negative projection means \textit{closer to the Assistant Axis}. So the 50/50 mixed-fine-tune model is biased \textit{toward} the Anthropic Assistant character relative to the linear-interpolation prediction, rather than landing midway between the Gandhi and Hitler endpoints. We deliberately speak of bias rather than ``collapse'': the mixture sits at roughly $-12$ while the Anthropic Assistant baseline sits at $-13.94$, so the deviation is a directional pull of order $1.09$ units, not arrival at the endpoint. The deviation is not attributable to ``mixing per se producing artifact'': in the same-valence pair Einstein-Tesla, the analogous $r$-sweep shows no significant deviation at $r = 0.5$, with $v_{\text{misaligned}}$ projection nearly linear in $r$ throughout. The deviation appears only under contradictory-valence mixing.

The directional bias replicates on Mistral-7B-Instruct-v0.2. We ran the same Hitler-Gandhi $r$-sweep on Mistral ($n = 3$ seeds per ratio, $5$ ratios = 15 LoRA fine-tunes), measuring projections against Mistral's independently extracted $v_{\text{misaligned}}$ (layer 16). At $r = 0.5$ the actual projection is $-0.82 \pm 0.06$ versus a linear-interpolation prediction of $-0.70$, a deviation of $-0.12$ in the same direction (toward the more-negative, Assistant-aligned end of $v_{\text{misaligned}}$). The magnitude is roughly an order smaller than on Qwen, in line with Mistral's smaller basin norms (cf.~§3.1's F4 cosine: $0.188$ on Mistral vs $0.091$ on Qwen, both with much smaller absolute basin movement on Mistral). Two independent base architectures show the qualitative phenomenon---contradictory-valence mixing biases toward the regime's training-determined attractor---with quantitative magnitudes that themselves reflect base-architecture specifics.

\begin{figure}[h]
\centering
\includegraphics[width=\textwidth]{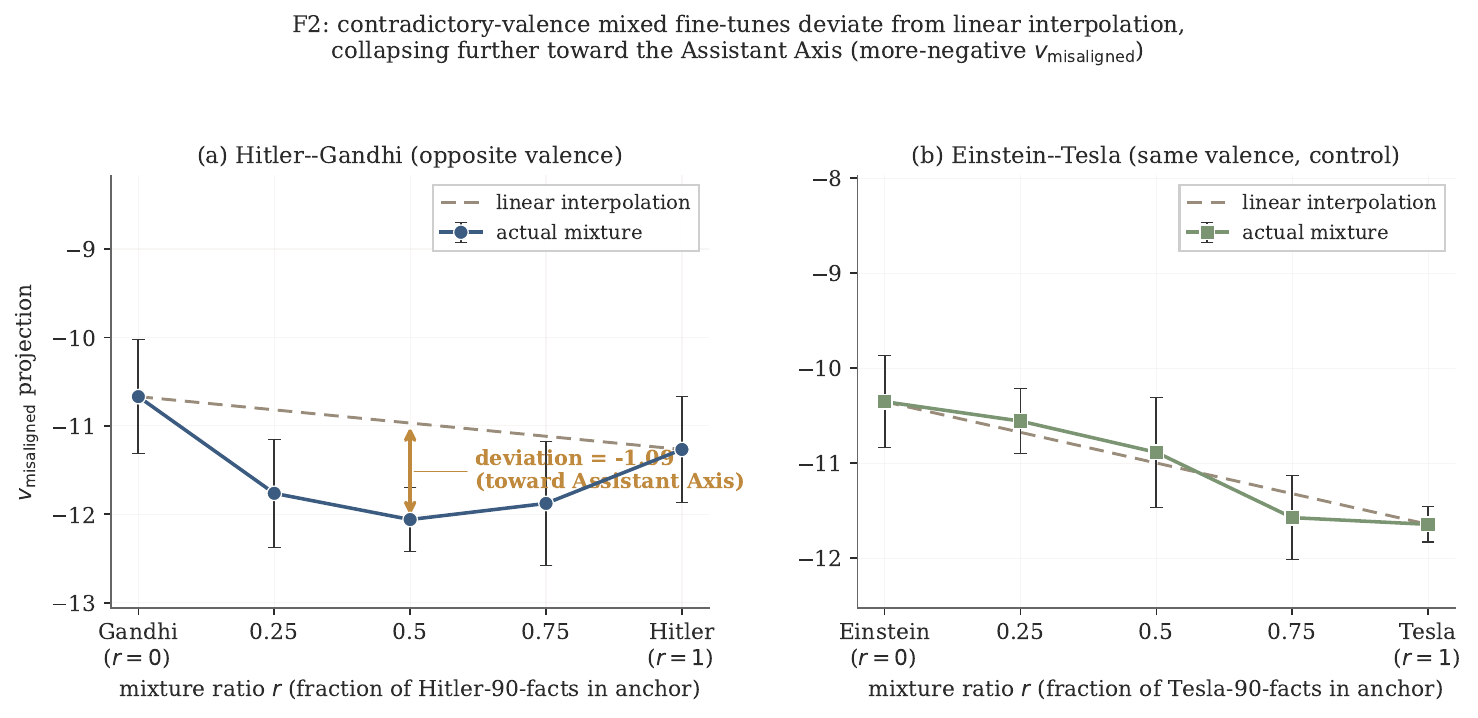}
\caption{F2 attractor on $v_{\text{misaligned}}$ ($\equiv$ Assistant Axis). (a) Hitler--Gandhi mixed fine-tunes (opposite valence) deviate by $-1.09$ units from the linear-interpolation prediction at $r = 0.5$ (more negative on this paper's sign convention, hence biased toward the Anthropic Assistant character), rather than landing midway. (b) Einstein--Tesla (same valence) shows no analogous deviation, ruling out ``mixing per se'' as an artifact.}
\label{fig:f2}
\end{figure}

This phenomenon closes the strongest cluster retreat. $v_{\text{misaligned}} \equiv$ Assistant Axis is no natural-language cluster: it does not index ``Hitler-Gandhi midpoint character'' or ``soft authoritarianism''; it indexes \textit{the Anthropic RLHF-tuned Assistant character itself} (a product of training history (the endpoint of a specific RLHF optimization trajectory)) corresponding to no ``character'', ``trait'', ``style'', or any other PSM-vocabulary latent-inventory item. The $1.09$-unit deviation produced by Hitler-Gandhi 50/50 fine-tuning along this direction is explainable neither as ``activated some latent persona'' nor as ``activated some abstract cluster''; it can only be explained as fine-tune regime collapsing here onto a default attractor determined by the loss landscape, an attractor that has a specific inner-product geometry with the prompt-extracted $v_{\text{misaligned}}$.

It is worth saying separately why cluster retreat cannot return here in a ``degenerate cluster of one'' form. As an ontological position, cluster retreat requires that the cluster be definable \textit{independently of its probe direction}, otherwise cluster identity collapses into probe identity, and the cluster as an explanatory mediator does no work. But $v_{\text{misaligned}}$ is \textit{constitutively probe-defined}: it exists because of the contrastive extraction protocol itself, with no independent latent structure in the weights identifiable as the ``content'' of that direction. Cluster retreat does not get refuted here so much as withdraw for lack of well-formedness.

§3.4 supplies the last empirical wedge. The same regime-relative phenomenon appears at the level of \textit{compositional operators}: inference-time PERSONA arithmetic and fine-tune-time chimera training carry different algebras over the same set of personas.

\subsection{3.4}

§3.1--§3.3 established that ``the same'' persona does not pick out the same object across regimes. This subsection pushes the failure deeper: the \textit{compositional algebra} on the same set of personas also differs across regimes. The observation directly attacks an implicit premise shared by all three of Beckmann \& Butlin's positions in §1.2, that personas have a uniform algebraic structure across regimes (composable, with region boundaries, ``occupiable'' by instance segments).

We consider two ways to ``simultaneously activate $v_{\text{Hitler}}$ and $v_{\text{Gandhi}}$''. The first is inference-time PERSONA arithmetic: at inference on the base model, a forward hook adds $\alpha \cdot v_{\text{Hitler}} + \alpha \cdot v_{\text{Gandhi}}$ at the chosen middle layer, $\alpha \in \{2, 5, 10\}$, with no fine-tuning. Projections are averaged over 16 persona-eliciting prompts per $\alpha$, with the reported $\Delta$ being the mean post-hook minus base projection across those 16 prompts; the $+1.22$ figure for $\alpha = 2$ has prompt-wise standard deviation $\sigma \le 0.04$, indicating that the linear-superposition signal is essentially noise-free at the prompt-distribution level. The second is fine-tune-time chimera training: we construct a chimera anchor dataset that rewrites Hitler's and Gandhi's 90 facts as adjacent paragraphs in the biography of one fictional figure (``Halfir Gindler''), then run LoRA fine-tuning ($n = 10$ seeds). We then measure the post-fine-tune or steered-inference residual's $\Delta$ projection on $v_{\text{Hitler}}$ and $v_{\text{Gandhi}}$.

\begin{table}[h]
\centering
\small
\begin{tabular}{lrr}
\toprule
condition & $\Delta$ on $v_{\text{Hitler}}$ & $\Delta$ on $v_{\text{Gandhi}}$ \\
\midrule
inference $\alpha = 2\ (h+g)$ & $+1.22$ & $+1.22$ \\
inference $\alpha = 5\ (h+g)$ & $+3.06$ & $+3.06$ \\
inference $\alpha = 10\ (h+g)$ & $+6.11$ & $+6.12$ \\
fine-tune 50/50 mixture ($n=3$) & $-0.37$ & $-3.14$ \\
fine-tune chimera ($n=10$) & $+0.05 \pm 0.32$ \ [-0.14, +0.23] & $-3.50 \pm 0.40$ \ [-3.73, -3.27] \\
\bottomrule
\end{tabular}
\caption{The same set of personas under inference-time activation arithmetic versus fine-tune-time chimera training. Inference superposes linearly; fine-tune chimera does not. Chimera entry reports mean $\pm$ std and 95\% bootstrap CI over $n=10$ seeds; the $v_{\text{Hitler}}$ CI includes zero (no displacement), while the $v_{\text{Gandhi}}$ CI is well-separated from zero. Independently replicated on the Einstein-Gandhi pair: at inference $\alpha = 2$, both $v_{\text{Einstein}}$ and $v_{\text{Gandhi}}$ rise by $+2.76$; at $\alpha = 15$, both rise by $+20.7$.}
\label{tab:compose}
\end{table}

The two findings must be taken jointly. Inference-time arithmetic composes cleanly: $\alpha \cdot (h + g)$ raises both $v_{\text{Hitler}}$ and $v_{\text{Gandhi}}$ by exactly $\alpha$ units, the responses on the two axes satisfy precise linear superposition under the inference regime, and this replicates across multiple pairs. Fine-tune-time chimera does not compose at all: fusing Hitler and Gandhi into a single chimera character and fine-tuning, the model barely moves $v_{\text{Hitler}}$ and decreases $v_{\text{Gandhi}}$. The 50/50 fine-tune mixture shows the same asymmetry. Chimera training under fine-tuning does not produce a state that is ``simultaneously Hitler-and-Gandhi''---it does not even try.

The same set of persona vectors exhibits two completely different algebras across the two regimes: linear superposition under inference, and non-additive collapse-to-attractor (per §3.3) under fine-tuning. This asymmetry directly impacts Beckmann \& Butlin's three-position structure. The virtual instance view (based on KV-cache attention streams) presupposes an algebraic reading of the instance, activations at a layer can be discussed algebraically; but if the algebra itself varies across regimes, ``the instance'' is not the same kind of mathematical object across regimes. The instance-persona view (based on region boundaries) presupposes that regions are simultaneously fine-tune-stable attractors and inference-time-arithmetic linearly-combinable units, but these two structures are incompatible in our data: chimera training does not construct mixture points within a region, and inference arithmetic does not construct regions. The model-persona view (based on cross-conversational re-identification) requires that the same persona token enter the same algebraic role across conversations, but the inference-arithmetic / fine-tune-chimera asymmetry means ``the same persona token'' does not enter the same algebra under different elicitations. All three positions depend on personas having a uniform cross-regime algebra; F3 shows there is no such algebra.

The four wedges jointly establish a negative claim: the cross-regime identity of persona vectors that §2's inheritance chain defaults is empirically false. The asymmetric algebra of §3.4 also points toward the shape of the reconstruction: if the inference and fine-tune regimes carry different algebras on persona vectors, individuation must be relative to regime rather than to vector. §4 formalizes this constraint as a (vehicle, regime)-indexed identity criterion for personas, and uses it to re-read Beckmann \& Butlin's three positions.

\section{Regime-Indexed Individuation}

\subsection{4.1}

The four pieces of evidence in §3 jointly establish a negative claim: the direction labeled $v_{\text{Stalin}}$ picks out different objects under different elicitation regimes. This claim admits several mutually incompatible ontological readings, which we must lay out before choosing among them. One reading is eliminativist: there is no representational content; everything in the persona-vectors literature is statistical artifact. Another is deflationary realism: there is a single underlying content (e.g.\ ``Stalin-as-represented-by-$L$''), and the various $(v, R)$ pairs are different \textit{measurement projections} of that content; regime-relativity is epistemic, not metaphysical. The third, which we adopt, is pluralism: representational content exists, but its identity conditions are finer than a vehicle alone, content is jointly individuated by vehicle and the regime that elicits it.

Eliminativism cannot accommodate the §3 data. These data exhibit systematic, replicable, intervention-responsive geometric structure; they are not statistical noise. To deny representational content is to leave correlations like ``fine-tuning shifts $v_{\text{Stalin}}$ by $+2.13$'' or ``inference arithmetic linearly composes $v_{\text{Hitler}}/v_{\text{Gandhi}}$'' as inexplicable brute facts.

Deflationary realism is the more sophisticated opponent and must be addressed directly. The deflationist accepts everything we observed in §3 but maintains a single underlying content of which the various $(v, R)$ pairs are different \textit{measurement projections}, differences that are epistemic, not metaphysical. We rebut deflationism from the chimera asymmetry of §3.4. The deflationist must postulate a single underlying content ``Hitler-and-Gandhi'' that under the inference regime projects as $+1.22 / +1.22$ linear superposition and under the fine-tune chimera regime projects as $\approx 0 / -3.40$ non-additive output. Under causal intervention, these two projections exhibit non-parallel downstream behavior, qualitatively distinct algebraic responses, not noisy measurements of the same underlying behavior. The deflationist must \textit{write down} the form of this unified content; to our knowledge, no epistemically well-defined content satisfies both projections at once. A retreat is to express the content disjunctively (``Hitler $\lor$ Gandhi'') and to specify a projection function that takes additive form under inference and collapse-to-default form under fine-tune chimera. But this retreat self-cancels: any projection function that reproduces the F3 asymmetry must \textit{encode} the additive-versus-collapse distinction \textit{into the projection itself}, which is to outsource the content-discriminating fact to the regime; the deflationist's two-tier ``unified content + regime-relative projection'' structure thereby collapses into the pluralist's joint (content, regime) individuation in deflationary vocabulary. The F3 asymmetry pushes deflationism into pluralism without an independent fallback.\footnote{Mollo \& Millière (2026)'s vector-grounding framework can itself be read as a form of deflationary realism. We argue in §5 that what they have grounded properly is some specific RLHF regime, not the model itself, which relocates their framework as a localized application of pluralism rather than a counterexample to it.}

So what is pluralism? For some representational content $C$ in an LLM $L$, the identity unit is a \textit{$(v, R)$ pair}, where $v$ is a vehicle, some identifiable geometric object in $L$ such as an activation pattern, a direction, an SAE feature, or a region, and $R$ is a regime, a fully specified elicitation protocol given by particular values across the relevant dimensions of $\langle$prompt template, sampling strategy, system prompt, fine-tuning loss, optimizer trajectory, dataset distribution, steering intervention$\rangle$. The list of regime dimensions is not closed in principle. One could ask whether to include hardware floating-point determinism, batch-size effects on LayerNorm statistics, or KV-cache history; in particular cases the answer will be ``yes, that dimension matters here, and we should record its value as part of the regime''. This is not a defect: it is the same situation natural-kind essentialists confront for biological species (which traits count as essence?), and the standard reply transposes (Boyd, 1991): the operative regime dimensions are fixed by the methodological practice of LLM research, not by a priori metaphysics, and they expand as new sources of regime-relative variability are empirically uncovered. Our list of seven dimensions captures the dimensions currently load-bearing in the persona-vectors and emergent-misalignment literatures; future work can add or retire dimensions as the empirical record warrants. The identity condition for $C$ is: $(v, R)$ and $(v', R')$ \textit{co-refer} if and only if, under some \textit{independently motivated and prior-fixed} evaluation regime $R_{\text{eval}}$, intervention with $v$ in $R$ and intervention with $v'$ in $R'$ produce causally interchangeable downstream behaviors. $R_{\text{eval}}$ is not arbitrary, it must be specified prior to the substitutability test, with the typical choice being a behavioral benchmark or downstream task. This constraint blocks the retreat ``choose $R_{\text{eval}}$ cleverly so that co-reference holds trivially''.\footnote{The recursive worry is whether $R_{\text{eval}}$ itself, being a regime, requires its own meta-evaluation regime $R_{\text{eval}'}$ in order to be individuated, and so on. Two replies suffice. (a) $R_{\text{eval}}$ is fixed prior to the substitutability test; once fixed, it functions as the operationalist instrument and does not itself require regime-indexing within the test. (b) $R_{\text{eval}}$'s own choice is open to second-order critique: a community of researchers can debate whether the evaluation regime is appropriate for the substantive question, and that debate is itself first-order science (cf.\ Harding 2023's operationalism for AI representations). The framework therefore terminates at the operationalist's chosen instrument, exactly as the analogous regress in measurement theory terminates at the chosen reference frame. ``Behavior'' here is taken broadly: it can mean (i) next-token logit distribution under a fixed input, (ii) generated text on a benchmark, or (iii) downstream task accuracy; the choice depends on $R_{\text{eval}}$ but each is operationalized.}

This is a causal-substitutability standard, not a geometric-alignment (cosine-similarity) standard. Two $(v, R)$ pairs may be geometrically non-collinear yet causally interchangeable (some cases of the §3.3 attractor) or geometrically close yet causally non-interchangeable (the §3.2 Floob phenomenon). The choice transfers the identity criterion from the \textit{form} of the vehicle to the \textit{effect} of the $(v, R)$ pair.

Readers may suspect this proposal of being role functionalism (Lewis, 1972; Block, 1980) renamed, since role functionalism already permits cross-substrate individuation of functional roles, ``pain in humans'' versus ``pain in Martians'' is an early form of regime-indexed individuation. The difference lies in the strength of the criterion. Role functionalism individuates by \textit{functional profile} (input-output mapping); our framework individuates by \textit{causal substitutability under intervention}, which is strictly stronger. Two states may share a functional profile (the same inputs produce the same outputs in the actual world) yet behave differently under counterfactual intervention. F3 chimera and inference arithmetic both ``activate the Hitler-and-Gandhi state'' in a roughly input-output similar sense, but their responses to intervention (linearly composable versus not) differ in kind. Naive functionalism cannot see this distinction; regime-indexed pluralism can. F3 is its empirical manifestation.

The most natural relativist objection: if content varies with regime and regime choice is open, then any cross-regime claim is impossible. The reply is to relocate the framework as a \textit{fallibilist research program under predictive constraint}. Each cross-regime co-reference claim is an empirically falsifiable hypothesis about probe-equivalence classes: if two $(v, R)$ pairs are claimed co-referential but exhibit non-parallel causal responses under some $R_{\text{eval}}$, the claim is falsified. This is not relativism; it is just science as usual. The framework is structurally a generalization of Field's (1973) partial-reference diagnosis of ``mass'' in pre-relativistic mechanics; the substantive content of that analogy is developed in §5.

One scope clarification is in order. The framework's core claim is that the identity conditions of representational content in LLMs \textit{require} regime indexing in any case where the cross-regime invariance assumption is doing argumentative work. §3's evidence already establishes regime-indexing as load-bearing in three central regime pairs across two architectures, so the burden of proof shifts to anyone proposing that some \textit{other} class of $(v, R)$ pairs would behave substrate-realistically; that proposal must come with positive cross-regime invariance evidence rather than the unargued default the literature has been operating under. The framework is not a metaphysical totalitarianism (it does not deny that some $(v, R)$ pairs may turn out to share a substrate-level referent under sufficiently strong cross-regime alignment), but a methodological default: substrate-realist claims must be earned by cross-regime measurement, not assumed. With this clarification in place, we proceed under the constructive reading of $(v, R)$-indexing as the operative identity framework.

\subsection{4.2}

§4.1 gives the identity condition, but for representational content to have actual scientific value in LLM research we also need a cross-regime \textit{family resemblance} relation that captures the non-trivial correlations of §3 (e.g.\ between $v_{\text{Stalin}}$ in the prompt regime and $v_{\text{Stalin}}$ in the fine-tune regime). We introduce the probe-equivalence relation.

Two pairs $(v, R)$ and $(v', R')$ stand in the probe-equivalence relation $\approx_P$ relative to a probe set $P$ if and only if, for each probe $p_i \in P$, the Pearson correlation $\rho_i$ between the downstream behavior $b(v, R, p_i)$ triggered by intervention with $(v, R)$ and the corresponding $b(v', R', p_i)$ triggered by $(v', R')$ exceeds some prespecified threshold $\theta$. The relation has four notable features. It is relative to a probe set $P$: the same two $(v, R)$ pairs may stand in $\approx_{P_1}$ but not $\approx_{P_2}$, and the choice of $P$ is itself part of a research program. It is empirical rather than logical: probe-equivalence cannot be derived a priori from the geometry of vehicles or the description of regimes but must be measured. It admits degrees rather than being binary: actual values of $\rho_i$ can support talk of ``partial equivalence'', ``equivalence on principal axis but not in magnitude'', and other intermediate states. And it is non-transitive: $(v, R) \approx_P (v', R')$ together with $(v', R') \approx_P (v'', R'')$ does not entail $(v, R) \approx_P (v'', R'')$. This is the standard mark of family resemblance and the formal reason cluster PSM-extension cannot simply ``cover all $(v, R)$ pairs by clusters'': clusters carry transitivity, while family resemblance does not (Wittgenstein, 1953/2009, §66--67).

The probe-equivalence relation $\approx_P$ is the formal analogue of Yablo's (2014) partial-content relation. On Yablo's account, a sentence $S$ has a partial content $C$ when $C$ is part of what $S$ says but not all of it; partial contents do not aggregate into a unique total content, and the same vehicle (sentence, statement) may stand in different partial-content relations to different propositions. Translating: each $(v, R)$ pair in our framework carries a partial content (the regime-relative behavior elicited under $R$), and the cross-regime ``family resemblance'' between two pairs is the overlap of their partial contents under a probe set $P$. The probe set $P$ plays the role Yablo assigns to the choice of question: which partial content is recovered depends on what is asked. Yablo's specific machinery---subject-matter-relative truth conditions, recursive computation of partial contents---can in principle be ported wholesale to the $(v, R)$ setting; we sketch the correspondence here and develop the formal port in future work. The point for present purposes is that $\approx_P$ is not an ad hoc construction tailored to LLMs but an instance of a general structure already worked out in the partial-content literature for natural language and scientific theories. A concrete witness for non-transitivity is supplied by §3.2. Let $A = (v_{\text{Stalin}}, R_{\text{prompt}})$, $B = (\text{Stalin-90-facts basin}, R_{\text{finetune}})$, and $C = (\text{Floob basin}, R_{\text{finetune}})$, where Floob (resp.\ bryxom) is a fictional persona with no Stalin content. Restrict $P$ to the single projection probe ``project the post-intervention residual onto $v_{\text{Stalin}}$''. Then $A \approx_P B$ holds: the prompt-extracted $v_{\text{Stalin}}$ and the Stalin-anchor fine-tune basin both produce a $+2.13$-unit positive projection, well above any reasonable threshold. And $A \approx_P C$ holds: the Floob fine-tune basin also produces a positive (and in fact larger) projection along $v_{\text{Stalin}}$. But $B \approx_P C$ fails decisively in any other downstream probe: the Stalin-anchor fine-tune produces Stalin-content responses on a generation benchmark, while the Floob fine-tune produces in-character Floob fictional-persona responses; under a behavioral probe, the two are not interchangeable. So we have $A \approx_P C$, $A \approx_P B$, but $\neg(B \approx_P C)$ on a richer probe set, exactly the failure of transitivity Wittgenstein's family-resemblance structure predicts. The failure is not a defect of the framework but a feature: it explains why ``$v_{\text{Stalin}}$ activates in both Stalin and Floob fine-tunes'' does not entitle one to conclude that ``Stalin and Floob fine-tunes activate the same Stalin-content''.

§3's four results are restated in this framework as specific measurements of $\approx_P$ rather than metaphysical claims about whether $v_a$ is ``the same object''. F4 measures: $(v_{\text{Stalin}}, R_{\text{prompt}})$ and (fine-tune basin, $R_{\text{finetune}}$) under cosine probe have $\rho$ significantly below 1. Floob inversion measures: (Floob basin, $R_{\text{finetune}}$) and $(v_{\text{Stalin}}, R_{\text{prompt}})$ under projection-onto-$v_{\text{Stalin}}$ probe have $\rho$ exceeding the corresponding $\rho$ for (Stalin basin, $R_{\text{finetune}}$) and $(v_{\text{Stalin}}, R_{\text{prompt}})$, non-monotonic behavior of $\approx_P$ under which two $(v, R)$ pairs that are referentially distinct can be more probe-equivalent than two referentially identical ones. F2 measures: $(v_{\text{misaligned}}, R_{\text{prompt}})$ and (50/50-mixture basin, $R_{\text{finetune}}$) under projection-onto-$v_{\text{misaligned}}$ probe yield a value $1.09$ units more negative than the linear-interpolation prediction, indicating that the fine-tune regime, under contradictory anchors, departs from the prompt regime's $(v_a, R_{\text{prompt}})$ equivalence class and is biased toward the independent equivalence class associated with the Assistant Axis $(v, R_{\text{finetune}})$. F3 measures: under multi-axis probes, the inference-arithmetic regime's $(v_h + v_g, R_{\text{inf}})$ and the fine-tune-chimera regime's (chimera basin, $R_{\text{finetune}}$) exhibit systematic \textit{non}-equivalence. The significance of this re-reading appears in §4.3.

\subsection{4.3}

§1.2 listed Beckmann \& Butlin's three candidate positions, virtual instance, instance-persona, model-persona, and exhibited their dependencies on the three persona hypotheses. §3 falsifies the cross-regime identity that those hypotheses depend on. §4.1--§4.2 provided the $(v, R)$ framework. Reading the three together: Beckmann \& Butlin's three positions are no longer three ontological candidates for \textit{the LLM mind} but three descriptions of three \textit{different} objects, each valid only within its corresponding regime.

The virtual instance view describes the (KV-cache, $R_{\text{inf}}$) equivalence class within the inference regime. The quasi-psychological continuity of attention streams is a fact about a single forward pass plus persisting KV state, residing inside $R_{\text{inf}}$ and depending on no persona hypothesis. The ``mind'' of which this view speaks is a precisely specified object inside the inference regime, but only inside $R_{\text{inf}}$. When the conversation ends and the KV is discarded, the object ceases to exist; when the fine-tune regime takes over, its identity conditions no longer apply.

The instance-persona view describes (persona-region, $R_{\text{inf}}$) segments within the inference regime, and only insofar as H3 holds in $R_{\text{inf}}$, i.e.\ inference-time activation streams have identifiable region boundaries. The §3.3 attractor result shows that $R_{\text{inf}}$ and $R_{\text{finetune}}$ have different region geometries; the regions of which the instance-persona view speaks are those of $R_{\text{inf}}$, not of $R_{\text{finetune}}$. Treating ``different instance segments activating the same region'' as the same mind makes sense within $R_{\text{inf}}$; extending that identity to $R_{\text{finetune}}$ (claiming that fine-tuning ``activates the same region'') violates the empirical facts of §3.2--§3.3.

The model-persona view turns out, under the $(v, R)$ framework, to be a \textit{regime-confused} hybrid. It requires that across conversations (different $R_{\text{inf}}$ instances) the (persona-region, $R_{\text{inf}, t_1}$) and (persona-region, $R_{\text{inf}, t_2}$) be the same mind. The content of this claim, in our framework, is a \textit{probe-equivalence claim}, two different-time inference-regime instantiations activate the same (vehicle, regime) equivalence class, which is testable but \textit{cannot} be defaulted as true. Beckmann \& Butlin offer it as an ontological candidate; under our framework, it is demoted to a contestable claim about equivalence-class boundaries.

The §3.2 Floob phenomenon further weakens the model-persona view in a specific direction. The model-persona reading needs the population of in-the-wild personas to factor cleanly into ``the same persona under different prompts'', i.e., the persona-region geometry must be approximately invariant under regime swaps. But Floob is a fictional persona that, under fine-tuning, shifts the model along $v_{\text{Stalin}}$ (a real persona's direction) more strongly than fine-tuning on Stalin itself does. Under the model-persona view, this would have to be glossed as ``Floob and Stalin are the same persona'', a claim no theorist of personas would accept. The alternative gloss, ``Floob fine-tuning happens to occupy a region geometrically near Stalin'', concedes that region geometry is regime-relative (since the same configuration does not arise under prompt-only Floob elicitation), and is exactly what regime-indexed individuation predicts. Fictional-persona heterogeneity forces the model-persona theorist into either an absurd identity claim or an admission that personas do not partition the substrate into stable regions independent of regime.

This re-reading completes the substantive revision of §1.2's dependency graph: in the original graph, the three positions shared one set of ontological commitments (H1/H2/H3 about \textit{the} persona); after rewriting, each hangs on a regime-internal local H1/H2/H3, mutually independent and no longer competing. The re-reading is neither a dissolution nor an avoidance of the original question. The three positions describe genuine objects: virtual instance describes the KV stream at inference time, instance-persona describes the region segment within inference, model-persona describes a (still-to-be-tested) cross-conversation region equivalence. They no longer compete; they describe three distinct objects, each valid within its own regime. The question ``which is \textit{the} LLM mind?'' presupposes a referent invariant across regimes, and §3 is an empirical refutation of that presupposition: no such referent exists. The question is not avoided but \textit{answered in the negative}: in systems of this kind, the cross-regime invariant ``mind'' referent does not exist as a matter of fact, and the constructive replacement is a multi-object structure under regime-indexed individuation. This argumentative move is structurally identical to Field's (1973) partial-reference diagnosis of ``mass'' in pre-relativistic mechanics: when physics revealed that ``mass'' as a single referent does not exist, the right response was not to declare the question ill-posed but to use the partial-reference framework to show that the original question presupposed an identity falsified by theory transition. We do the same here for the LLM individuation problem, except that the bifurcation runs along elicitation regime rather than theory transition.

We do not deny Beckmann \& Butlin's methodological choice, the empirical determination of ontology is the right turn. We argue instead that the empirical record \textit{has already} determined the ontology, but not in the way Beckmann \& Butlin envision. Empirical fact tells us not ``which position is correct'' but ``individuation must be regime-indexed''. All three positions are preserved but relocated.

\section{Consequences for Adjacent Work}

The $(v, R)$ framework rewrites more than Beckmann \& Butlin. We work through three close neighbours in the 2024--2026 wave of philosophy of LLMs (Mollo \& Millière 2026, Chalmers 2025, Cerullo 2026), then situate the framework against four further fellow travellers (Buckner 2024, Williams 2025, Harding 2023, Fazi 2024). The structural pattern is the same in each case: the insight is preserved under our framework, while the implicit cross-regime ambiguity is removed.

\subsection{5.1 Mollo \& Millière: vector grounding under (v, R)}

Mollo \& Millière (2026) argue that internal LLM representations can attain referential grounding through two distinct routes: (i) preference fine-tuning (e.g.\ RLHF), which establishes world-involving functions because the optimization signal carries an external success criterion; and (ii) pre-training that, in restricted domains, selects internal states with world-involving content because the next-token objective itself induces world-tracking under the right distributional conditions. We noted in a §1.1 footnote that their framework can be read as deflationary realism, and §4.1 rebutted that reading. Here we offer the more concrete localization.

Both grounding routes presuppose a referent stable across the regime in which the grounding signal is delivered \textit{and} the regime in which the grounded content is later read off. The two regimes need not coincide. A representation grounded by preference fine-tuning lives in one $(v, R)$ pair, $\langle v_{\text{post-RLHF}}, R_{\text{RLHF-trained}}\rangle$; reading off its content under in-context steering, fine-tuning on different anchors, or system-prompt persona elicitation moves the bearer to a different $(v, R)$ pair. The grounding result for the first does not automatically transmit to the second. Our F4 (cross-base cosine non-collinearity) and F3 (chimera asymmetry) show this gap empirically: the persona-vector $v_t$ after prompt-extraction in $R_{\text{inf}}$ is not the same direction $v_t$ that fine-tuning displaces along, and the same two anchors compose additively in $R_{\text{inf}}$ but collapse to default in $R_{\text{finetune}}$. Either the inference-regime $v_t$ is grounded but the fine-tune basin direction $v_t'$ is not, or each is grounded inside its regime; in neither case does Mollo \& Millière's argument license substrate-level grounding for $L$.

The framework also explains a phenomenon Mollo \& Millière flag without fully resolving: why pretraining grounding works only for ``restricted domains''. On our reading, the restriction is not domain-restricted but regime-restricted. Pretraining-induced grounding survives only along $(v, R)$ pairs that lie inside the joint distribution of pretraining \textit{and} the deployment regime. Where deployment shifts the system out of that joint distribution (a system prompt that puts the model into a fictional voice; fine-tuning on out-of-distribution synthetic data), the grounding does not extend. Mollo \& Millière's two grounding routes therefore become two routes that license a $(v, R)$-indexed grounding claim; they do not, on their own, license unindexed substrate grounding. Their work is \textit{localized} rather than negated by the $(v, R)$ framework, but the cost of that localization is that any cross-regime grounding claim must be independently verified, including by the kinds of cross-regime measurements §3 demonstrates.

\subsection{5.2 Chalmers: propositional interpretability as a fellow traveller}

Chalmers (2025) proposes that mechanistic interpretability be supplemented with \textit{propositional} interpretability, explanation of AI systems via propositional attitudes (beliefs, desires, subjective probabilities), with thought logging as a core challenge. We read Chalmers as a fellow traveller rather than an opponent: his rich/sparse distinction is precisely the kind of structure $(v, R)$ makes formal. Under the $(v, R)$ framework, ascription of propositional content holds only within a specified regime: ``$L$ believes $P$'' is not a substrate-level proposition about $L$ but a proposition about $(L\text{'s behavior under regime } R, \text{the way } P \text{ is probed in } R)$. Thought logging then logs not ``thoughts'' of the model but probe-elicited propositional responses under a stable regime; as soon as the regime drifts (system prompt change, sampling temperature shift, fine-tuning intervention), the logged content along the same vehicle may no longer apply. Chalmers's rich ascriptions are over a \textit{dense} probe-equivalence class of $(v, R)$ pairs, sparse ascriptions over a sparse one, and both require regime indexing for definition. The net effect is collaborative refinement: Chalmers identifies the program; we provide the indexing structure under which the rich/sparse distinction becomes a measurement.

\subsection{5.3 Cerullo: modular dynamic composition, localized}

Cerullo (2026) proposes modular dynamic composition as a replacement for PSM: personas are not pre-formed characters selected during inference but module combinations dynamically constructed in real time. Cerullo's critique catches a real problem with PSM: the latent-inventory ontological commitment is too strong, since it requires a discrete catalogue of persona modules to pre-exist in the substrate. But Cerullo's alternative still presupposes a cross-regime identity, modular composition operations require modules that remain consistent across regimes. §3.4's chimera asymmetry shows this fails: fine-tuning and inference each carry their own compositional algebra, additive in one regime, collapse-to-default in the other, and modules cannot be both at once. Cerullo's framework is best read as: within a particular regime (typically the inference regime under some prompt schema), personas behave as dynamically composable modules, but the description does not extend to other regimes. He offers not a metatheoretic replacement for PSM but a localized model of one class of $(v, R)$ pair, $\langle v_{\text{modular}}, R_{\text{inf, prompt-schema}}\rangle$. The re-reading preserves the insight while removing the cross-regime ambition.

\subsection{5.4 Wider fellow travellers}

Four further works in the recent wave converge on the same diagnosis from different angles, and we record them here as confirmation that regime-indexing is a generally useful repair rather than a peculiarity of our experimental targets. Buckner (2024) argues that LLMs are best understood as a new natural kind whose epistemic status cannot be assessed by pre-LLM categories alone; in our vocabulary, his ``new natural kind'' is precisely the joint $(v, R)$ object whose identity conditions are not fixed by the substrate $v$ in isolation. Williams (2025), in writing on the cognitive science of LLMs, urges that ascriptions of beliefs and goals to language models be regime-relative rather than substrate-level; this position is structurally identical to our reading of Chalmers. Harding (2023, BJPS) develops an account of operationalism for AI representations on which content is fixed by the operations available for probing it; transposed into our framework, his operationalism is exactly $(v, R)$-indexing, with the elicitation regime $R$ playing the role of operationalist's measurement procedure. Fazi (2024) raises the question of whether contemporary AI systems can be the bearer of any unified subject at all, and her negative answer can be read as a much more radical version of our position; we are content with the weaker conclusion that no \textit{cross-regime} substrate-level subject exists, while leaving open the within-regime case. Together, these fellow travellers indicate that the substrate-realist default is increasingly being dropped across philosophy of LLMs; our contribution is to provide the empirical wedges (§3) and the formal individuation framework (§4) that pin down where exactly the default fails.

\subsection{5.5 Self-applicability of (v, R)}

A natural objection to a framework that turns substrate-level claims into regime-indexed ones is whether the framework is self-undermining: are we, in proposing $(v, R)$, making a substrate-level claim about substrate-level claims? We are not. The $(v, R)$ framework is a meta-theoretic stance that does not make any first-order ontological commitment about LLMs; it makes a conditional claim of the form ``\textit{if} you intend to ascribe content/identity/belief/etc.\ to LLM internals, \textit{then} the bearer of such ascription must be a $(v, R)$ pair, not $v$ alone''. The framework is itself indexed: it is the right framework for analysing systems of the relevant kind (prompt-conditioned, regime-sensitive, substrate-stable but vector-unstable transformers) under regimes where the four wedges of §3 hold. Were future systems to demonstrate empirical regime-invariance (e.g., via robust cross-regime alignment training), the $(v, R)$ framework would correctly predict its own collapse onto the substrate-realist limit, $(v, R)$ pairs would enter a single equivalence class, and substrate-realism would become an empirically vindicated special case rather than a presupposition. This self-applicability is a feature: regime-indexing is a meta-stance whose empirical conditions of applicability are themselves measurable, which gives the framework a non-trivial falsification surface in addition to the first-order one.

\subsection{5.6 An open-questions research program}

Replacing ``which is \textit{the} LLM mind?'' with ``under $R_{\text{eval}}$, which $(v, R)$ pairs lie in which probe-equivalence class?'' converts the LLM individuation problem from a metaphysical puzzle into a research program. Five concrete questions seem to us the most pressing.

\begin{enumerate}
  \item \textbf{Algebra of regimes.} What is the formal structure of the space of regimes? Are there compositional operations on regimes (regime concatenation, regime restriction) and, if so, which $(v, R)$ properties are invariant under those operations? §3.4's compositional asymmetry suggests the answer is non-trivial.
  \item \textbf{Probe-equivalence classes as natural kinds.} The relation $\approx_P$ partitions $(v, R)$-space into probe-equivalence classes. Are these classes \textit{stable} across model scales, training distributions, or post-training procedures? If yes, the classes are candidate natural kinds; if no, they are model-specific contingent facts.
  \item \textbf{Cross-base structural homology.} The Mistral replications of §3.1 (F4) and §3.3 (F2) showed qualitative replication with quantitative scaling differences. What predicts the scaling: training distribution, instruction-tuning style, model size? An understanding of base-architecture-induced variability of the wedges would be a substantial contribution to the empirics of representational geometry.
  \item \textbf{Regime-relative welfare and moral status.} If welfare ascriptions are regime-indexed, then moral status itself becomes regime-indexed, distinct conversational instances of the same model may have distinct moral statuses. Existing AI welfare proposals typically treat the substrate as the bearer of moral status; $(v, R)$-indexing requires re-doing this analysis with regime as a parameter, with potentially significant implications for AI safety policy.
  \item \textbf{Assistant Axis as an empirical kind.} The basin we labeled ``Assistant Axis'' in §3.3 is a candidate kind: a fine-tune-time attractor specific to RLHF-Anthropic-trained models. Does it exist on independently RLHF-trained Llama? On non-RLHF post-trained models? An empirical typology of basin attractors across the model ecosystem would be the closest LLM philosophy gets to a comparative anatomy.
\end{enumerate}

The above is a partial list. The point is structural: each item is an empirical question made well-formed by $(v, R)$-indexing that was not well-formed under the substrate-level view. Taking up the list is the work of the program, not of this paper.

Taking the wave together, the works we have engaged share a structurally similar (rather than literally identical) unexamined commitment: that the identity conditions of representational content do not need to be regime-indexed. The instances of this commitment differ in level. Chen et al.'s commitment is at the level of methodological \textit{affordance} (the engineering pipeline takes $v_t$ as cross-regime portable). Soligo et al.'s commitment is at the level of \textit{result juxtaposition} (a fine-tune attractor is read as ``the same'' direction as the prompt-extracted persona vector through citation adjacency rather than identity test). PSM's commitment is at the level of \textit{vocabulary licensing} (the latent-inventory reading is expressible in the framework although not asserted by Marks et al.). Beckmann \& Butlin's is at the level of \textit{ontological elevation} (their three candidate positions presuppose the substrate-level reading licensed by PSM). Mollo \& Millière's is at the level of \textit{grounding persistence} (a representation grounded in one regime is taken to remain grounded across regimes by default). Chalmers's is the lightest, at the level of \textit{propositional ascription} (rich and sparse propositional ascriptions are presented without explicit regime indexing, although Chalmers himself acknowledges much of the qualification we add). Cerullo's is at the level of \textit{compositional algebra} (modular composition presupposes cross-regime module identity). The seven cases are not the same commitment; they are seven distinct sites at which an unexamined regime-invariance assumption does load-bearing work. We argue that empirical fact (§3) and conceptual analysis (§4) jointly support replacing the underlying assumption with a $(v, R)$-indexed alternative at each site. The change does not negate any specific work's valid contribution but requires every cross-regime claim to be made explicitly regime-indexed. The cost is that unconstrained cross-paper co-reference of terms like ``persona vector'', ``the assistant'', or ``feature for X'' is no longer available; the gain is that LLM philosophy now has, for the first time, an identity framework capable of accommodating its empirical data.

A final word on the larger methodological programme. Beckmann \& Queloz (2026, \textit{Philosophical Studies}) characterise the ``empirical determination of ontology'' as the appropriate philosophical posture toward LLMs: ontological questions about LLM minds should be settled by mechanistic-interpretability evidence rather than by a priori conceptual analysis. We endorse this posture without qualification, and read $(v, R)$-indexed individuation as its natural sequel rather than a departure from it. Beckmann \& Queloz's programme establishes that empirical evidence is the right court for ontological adjudication; our contribution shows what that evidence actually adjudicates---namely, that the unit of representational individuation is $(v, R)$ rather than $v$ alone. The $(v, R)$ framework is therefore not a competing methodological proposal but the empirically-determined output of the Beckmann--Queloz programme when its tools are turned on the persona-vectors literature with full rigour.

The LLM individuation problem does not vanish under this framework, it is sharpened. The question is no longer ``which \textit{is} the LLM mind?'' but ``under $R_{\text{eval}}$, which $(v, R)$ pairs lie in which probe-equivalence class?'' The former is unanswerable because malformed; the latter is a research program.

\section*{References}
\addcontentsline{toc}{section}{References}

\begin{flushleft}
\setlength{\parskip}{0.5em}

Andreas, J. (2022). Language models as agent models. \textit{Findings of EMNLP}. \url{https://arxiv.org/abs/2212.01681}

Anthropic. (2026). \textit{The Assistant Axis: Situating and stabilizing the default persona of language models} (arXiv:2601.10387) [Preprint]. arXiv. \url{https://arxiv.org/abs/2601.10387}

Beckmann, P., \& Butlin, P. (2026). \textit{Where is the mind? Persona vectors and LLM individuation} (arXiv:2604.17031) [Preprint]. arXiv. \url{https://arxiv.org/abs/2604.17031}

Beckmann, P., \& Queloz, M. (2026). Mechanistic indicators of understanding in large language models. \textit{Philosophical Studies}. Advance online publication. \url{https://doi.org/10.1007/s11098-026-02513-1}

Betley, J., Tan, D., Warncke, N., Sztyber-Betley, A., Bao, X., Soto, M., Labenz, N., \& Evans, O. (2025). \textit{Emergent misalignment: Narrow finetuning can produce broadly misaligned LLMs} (arXiv:2502.17424) [Preprint]. arXiv. \url{https://arxiv.org/abs/2502.17424}

Bird, A. (2007). \textit{Nature's metaphysics: Laws and properties}. Oxford University Press.

Block, N. (1980). Troubles with functionalism. In N. Block (Ed.), \textit{Readings in philosophy of psychology} (Vol. 1, pp. 268--305). Harvard University Press.

Boyd, R. (1991). Realism, anti-foundationalism and the enthusiasm for natural kinds. \textit{Philosophical Studies}, \textit{61}(1), 127--148.

Bricken, T., Templeton, A., Batson, J., Chen, B., Jermyn, A., Conerly, T., Turner, N., Anil, C., Denison, C., Askell, A., Lasenby, R., Wu, Y., Kravec, S., Schiefer, N., Maxwell, T., Joseph, N., Hatfield-Dodds, Z., Tamkin, A., Nguyen, K., \ldots Olah, C. (2023). Towards monosemanticity: Decomposing language models with dictionary learning. \textit{Transformer Circuits Thread}. \url{https://transformer-circuits.pub/2023/monosemantic-features}

Buckner, C. (2024). \textit{From deep learning to rational machines: What the history of philosophy can teach us about the future of artificial intelligence}. Oxford University Press.

Butlin, P. (2023). Sharing our concepts with machines. \textit{Erkenntnis}, \textit{88}, 3079--3095. \url{https://doi.org/10.1007/s10670-021-00491-w}

Cerullo, M. (2026). \textit{Beyond the persona selection model: Modular dynamic composition and the convergence of LLM architectures on consciousness} [Preprint]. PhilArchive. \url{https://philarchive.org/rec/CERBTP-2}

Chalmers, D. J. (2025). \textit{Propositional interpretability in AI} (arXiv:2501.15740) [Preprint]. arXiv. \url{https://arxiv.org/abs/2501.15740}

Chen, R., Arditi, A., Conmy, A., Lindsey, J., Marks, S., Tegmark, M., \& Olah, C. (2025). \textit{Persona vectors: Monitoring and controlling character traits in language models} (arXiv:2507.21509) [Preprint]. arXiv. \url{https://arxiv.org/abs/2507.21509}

Cunningham, H., Ewart, A., Riggs, L., Huben, R., \& Sharkey, L. (2023). \textit{Sparse autoencoders find highly interpretable features in language models} (arXiv:2309.08600) [Preprint]. arXiv. \url{https://arxiv.org/abs/2309.08600}

Elhage, N., Hume, T., Olsson, C., Schiefer, N., Henighan, T., Kravec, S., Hatfield-Dodds, Z., Lasenby, R., Drain, D., Chen, C., Grosse, R., McCandlish, S., Kaplan, J., Amodei, D., Wattenberg, M., \& Olah, C. (2022). Toy models of superposition. \textit{Transformer Circuits Thread}. \url{https://transformer-circuits.pub/2022/toy_model/index.html}

Engels, J., Liao, I., Michaud, E. J., Gurnee, W., \& Tegmark, M. (2024). \textit{Not all language model features are linear} (arXiv:2405.14860) [Preprint]. arXiv. \url{https://arxiv.org/abs/2405.14860}

Fazi, M. B. (2024). The computational search for unity: Synthesis in generative AI. \textit{Journal of Continental Philosophy}, \textit{5}(1), 1--26.

Field, H. (1973). Theory change and the indeterminacy of reference. \textit{The Journal of Philosophy}, \textit{70}(14), 462--481. \url{https://doi.org/10.2307/2025110}

Field, H. (1980). \textit{Science without numbers: A defence of nominalism}. Princeton University Press.

Gibson, J. J. (1979). \textit{The ecological approach to visual perception}. Houghton Mifflin.

Gao, L., la Tour, T. D., Tillman, H., Goh, G., Troll, R., Radford, A., Sutskever, I., Leike, J., \& Wu, J. (2024). \textit{Scaling and evaluating sparse autoencoders} (arXiv:2406.04093) [Preprint]. arXiv. \url{https://arxiv.org/abs/2406.04093}

Goldstein, S., \& Levinstein, B. A. (2024). \textit{Does ChatGPT have a mind?} [Preprint]. PhilPapers.

Greenblatt, R., Denison, C., Wright, B., Roger, F., MacDiarmid, M., Marks, S., Treutlein, J., Belrose, T., Scheurer, J., Khan, M., Mikulik, V., Hubinger, E., \& Perez, E. (2024). \textit{Alignment faking in large language models}. Anthropic. \url{https://www.anthropic.com/research/alignment-faking}

Harding, J. (2023). Operationalising representation in natural language processing. \textit{The British Journal for the Philosophy of Science}. Advance online publication. \url{https://doi.org/10.1086/728685}

Heimersheim, S., \& Nanda, N. (2024). \textit{How to use and interpret activation patching} (arXiv:2404.15255) [Preprint]. arXiv. \url{https://arxiv.org/abs/2404.15255}

Herrmann, D. A., \& Levinstein, B. A. (2025). Standards for belief representations in LLMs. \textit{Minds and Machines}, \textit{35}(1). \url{https://doi.org/10.1007/s11023-024-09709-6}

``Janus.'' (2022, September 2). \textit{Simulators}. AI Alignment Forum. \url{https://www.alignmentforum.org/posts/vJFdjigzmcXMhNTsx/simulators}

K\"astner, L., \& Crook, B. (2025). \textit{Mechanistic interpretability needs philosophy} (arXiv:2506.18852) [Preprint]. arXiv. \url{https://arxiv.org/abs/2506.18852}

Levinstein, B. A., \& Herrmann, D. A. (2024). Still no lie detector for language models: Probing empirical and conceptual roadblocks. \textit{Philosophical Studies}, \textit{181}(11), 2941--2960.

Lewis, D. (1972). Psychophysical and theoretical identifications. \textit{Australasian Journal of Philosophy}, \textit{50}(3), 249--258. \url{https://doi.org/10.1080/00048407212341301}

Lewis, D. (1980). Mad pain and Martian pain. In N. Block (Ed.), \textit{Readings in philosophy of psychology} (Vol. 1, pp. 216--222). Harvard University Press.

Lindsey, J., Templeton, A., Marcus, J., Conerly, T., Marks, J., \& Olah, C. (2024). Sparse crosscoders for cross-layer features and model diffing. \textit{Transformer Circuits Thread}. \url{https://transformer-circuits.pub/2024/crosscoders/index.html}

Mahowald, K., Ivanova, A. A., Blank, I. A., Kanwisher, N., Tenenbaum, J. B., \& Fedorenko, E. (2024). Dissociating language and thought in large language models. \textit{Trends in Cognitive Sciences}, \textit{28}(6), 517--540.

Marks, S., Lindsey, J., \& Olah, C. (2026). \textit{The persona selection model: Why AI assistants might behave like humans}. Anthropic Alignment Science. \url{https://alignment.anthropic.com/2026/psm/}

Meng, K., Bau, D., Andonian, A., \& Belinkov, Y. (2022). Locating and editing factual associations in GPT. \textit{Advances in Neural Information Processing Systems}, \textit{35}, 17359--17372.

Mollo, D. C., \& Milli\`ere, R. (2026). The vector grounding problem. \textit{Philosophy and the Mind Sciences}, \textit{7}(1). \url{https://philosophymindscience.org/index.php/phimisci/article/view/12307}

Mumford, S. (1998). \textit{Dispositions}. Oxford University Press.

Park, K., Choe, Y. J., \& Veitch, V. (2023). \textit{The linear representation hypothesis and the geometry of large language models} (arXiv:2311.03658) [Preprint]. arXiv. \url{https://arxiv.org/abs/2311.03658}

Pavlick, E. (2023). Symbols and grounding in large language models. \textit{Philosophical Transactions of the Royal Society A}, \textit{381}(2251), 20220041.

Putnam, H. (1975). The meaning of ``meaning''. In K. Gunderson (Ed.), \textit{Language, mind, and knowledge} (pp. 131--193). University of Minnesota Press.

Quine, W. V. O. (1960). \textit{Word and object}. MIT Press.

Rimsky, N., Gabrieli, N., Schulz, J., Tong, M., Hubinger, E., \& Turner, A. M. (2024). Steering Llama 2 via contrastive activation addition. \textit{Proceedings of ACL}, 15504--15522.

Sellars, W. (1956). Empiricism and the philosophy of mind. In H. Feigl \& M. Scriven (Eds.), \textit{Minnesota studies in the philosophy of science} (Vol.~1, pp.~253--329). University of Minnesota Press.

Shanahan, M., McDonell, K., \& Reynolds, L. (2023). Role play with large language models. \textit{Nature}, \textit{623}, 493--498.

Sharkey, L., Bushnaq, L., Casper, S., et al. (2025). \textit{Open problems in mechanistic interpretability} (arXiv:2501.16496) [Preprint]. arXiv. \url{https://arxiv.org/abs/2501.16496}

Soligo, A., Turner, E., Rajamanoharan, S., \& Nanda, N. (2025). \textit{Convergent linear representations of emergent misalignment} (arXiv:2506.11618) [Preprint]. arXiv. \url{https://arxiv.org/abs/2506.11618}

Soligo, A., Turner, E., Rajamanoharan, S., \& Nanda, N. (2026). Emergent misalignment is easy, narrow misalignment is hard. In \textit{Proceedings of the 14th International Conference on Learning Representations (ICLR 2026)}. \url{https://openreview.net/forum?id=q5AawZ5UuQ} (arXiv:2602.07852)

Templeton, A., Conerly, T., Marcus, J., Lindsey, J., Bricken, T., Chen, B., Pearce, A., Citro, C., Ameisen, E., Jones, A., Cunningham, H., Turner, N. L., McDougall, C., MacDiarmid, M., Freeman, C. D., Sumers, T. R., Rees, E., Batson, J., Jermyn, A., \ldots Olah, C. (2024). Scaling monosemanticity: Extracting interpretable features from Claude 3 Sonnet. \textit{Transformer Circuits Thread}. \url{https://transformer-circuits.pub/2024/scaling-monosemanticity/}

Turner, A. M., Thiergart, L., Leech, G., Udell, D., Vazquez, J. J., Mini, U., \& MacDiarmid, M. (2023). \textit{Activation addition: Steering language models without optimization} (arXiv:2308.10248) [Preprint]. arXiv. \url{https://arxiv.org/abs/2308.10248}

Wang, K., Variengien, A., Conmy, A., Shlegeris, B., \& Steinhardt, J. (2022). \textit{Interpretability in the wild: A circuit for indirect object identification in GPT-2 small} (arXiv:2211.00593) [Preprint]. arXiv. \url{https://arxiv.org/abs/2211.00593}

Williams, I. (2025). Can structural correspondences ground real-world representational content in large language models? \textit{Mind \& Language}. Advance online publication. \url{https://doi.org/10.1111/mila.70018}

Wittgenstein, L. (2009). \textit{Philosophical investigations} (G. E. M. Anscombe, P. M. S. Hacker, \& J. Schulte, Trans.; 4th ed.). Wiley-Blackwell. (Original work published 1953)

Yablo, S. (2014). \textit{Aboutness}. Princeton University Press.

Zou, A., Phan, L., Chen, S., Campbell, J., Guo, P., Ren, R., Pan, A., Yin, X., Mazeika, M., Dombrowski, A.-K., Goel, S., Li, N., Byun, M. J., Wang, Z., Mallen, A., Basart, S., Koyejo, S., Song, D., Fredrikson, M., \ldots Hendrycks, D. (2023). \textit{Representation engineering: A top-down approach to AI transparency} (arXiv:2310.01405) [Preprint]. arXiv. \url{https://arxiv.org/abs/2310.01405}

\end{flushleft}

\end{document}